\pdfoutput=1
\documentclass[11pt]{article}

\usepackage{ACL2023}

\usepackage{times}
\usepackage{latexsym}

\usepackage[T1]{fontenc}
\usepackage[utf8]{inputenc}

\usepackage{microtype}

\usepackage{inconsolata}

\usepackage{booktabs}
\usepackage{comment}
\usepackage{amsmath,amssymb,amsthm}
\usepackage{url}
\usepackage{multirow}
\usepackage{graphicx}
\usepackage{subfigure}
\usepackage{siunitx,booktabs}
\usepackage{tabularx,comment}
\usepackage{makecell}
\usepackage{arydshln}
\usepackage{multirow}
\usepackage{float}
\usepackage{xspace}
\usepackage[italian,arabic,farsi,english]{babel}
\usepackage{cjhebrew}
\usepackage{devanagari}
\usepackage{CJKutf8}
\usepackage{hyphenat}
\usepackage{color}
\usepackage{pifont}
\newcommand{\tick}{\ding{51}}
\newcommand{\cross}{\ding{55}}
\usepackage{listings}
\usepackage{microtype}
\usepackage{enumitem}
\usepackage{setspace}

\newcommand{\ours}{\textsc{XSemPLR}\xspace}

\title{\ours: Cross-Lingual Semantic Parsing in Multiple Natural Languages and Meaning Representations}
\author{
Yusen Zhang$^1$ 
\quad Jun Wang$^2$ 
\quad Zhiguo Wang$^2$
\quad Rui Zhang$^1$
\\
$^1$Penn State University \quad $^2$AWS AI Labs
\\
\tt{\{yfz5488,rmz5227\}@psu.edu},
\tt{\{juwanga,zhiguow\}@amazon.com}
}

\begin{document}
\maketitle
\begin{abstract}

Cross-Lingual Semantic Parsing (CLSP) aims to translate queries in multiple natural languages (NLs) into meaning representations (MRs) such as SQL, lambda calculus, and logic forms.
However, existing CLSP models are separately proposed and evaluated on datasets of limited tasks and applications, impeding a comprehensive and unified evaluation of CLSP on a diverse range of NLs and MRs.
To this end, we present \ours, a unified benchmark for cross-lingual semantic parsing featured with 22 natural languages and 8 meaning representations by examining and selecting 9 existing datasets to cover 5 tasks and 164 domains. 
We use \ours to conduct a comprehensive benchmark study on a wide range of multilingual language models including encoder-based models (mBERT, XLM-R), encoder-decoder models (mBART, mT5), and decoder-based models (Codex, BLOOM).
We design 6 experiment settings covering various lingual combinations (monolingual, multilingual, cross-lingual) and numbers of learning samples (full dataset, few-shot, and zero-shot).
Our experiments show that encoder-decoder models (mT5) achieve the highest performance compared with other popular models, and multilingual training can further improve the average performance.
Notably, multilingual large language models (e.g., BLOOM) are still inadequate to perform CLSP tasks.
We also find that the performance gap between monolingual training and cross-lingual transfer learning is still significant for multilingual models, though it can be mitigated by cross-lingual few-shot training.
Our dataset and code are available at \url{https://github.com/psunlpgroup/XSemPLR}.
\end{abstract}

\section{Introduction}

\begin{figure*}[!t]
    \centering
    \includegraphics[width=\linewidth]{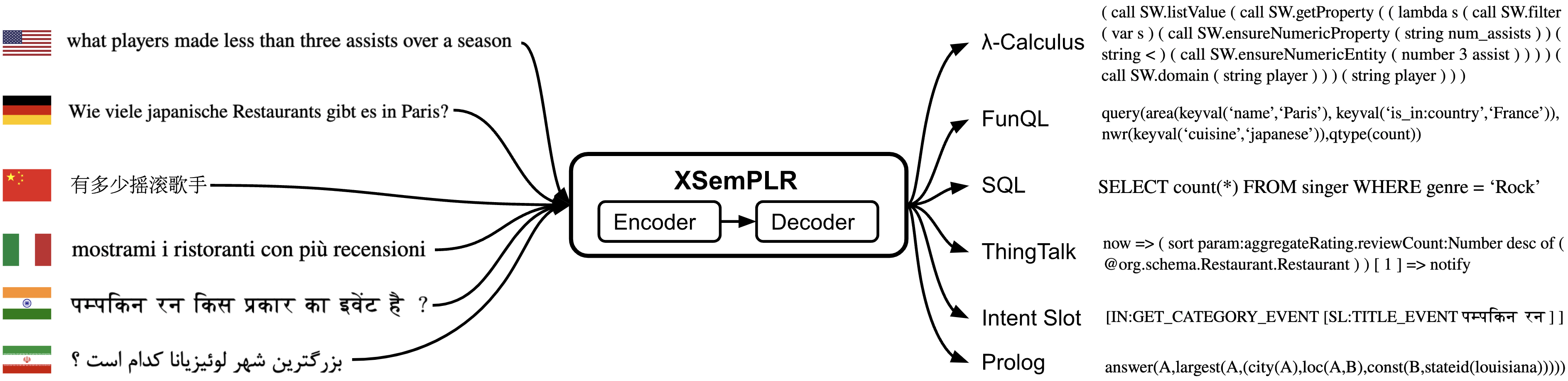}
    \caption{Overview of Cross-Lingual Semantic Parsing over various natural languages and meaning representations.}
    \label{fig:clsp}
\end{figure*}

Cross-Lingual Semantic Parsing (CLSP) aims to translate queries in multiple natural languages (NLs) into meaning representations (MRs)~\citep{li2020mtop,xu2020schema2qa,dou2022multispider,sherborne2021zero,sherborne-lapata-2022-zero}.
As demonstrated in Figure~\ref{fig:clsp}, Cross-Lingual Semantic Parsing covers natural languages for geographically diverse users and various meaning representations, empowering applications such as natural language interfaces to databases, question answering over knowledge graphs, virtual assistants, smart home device control, human-robot interaction, and code generation.

However, current research on CLSP has three drawbacks.
First, most existing research focuses on semantic parsing in English~\cite{zelle1996learning,wang2015building,yu2018spider}, limiting the development of multilingual information access systems for users in other languages.
Second, current datasets have a poor coverage of NLs and MRs. Although there are encouraging efforts in developing CLSP models~\citep{li2020mtop,dou2022multispider,sherborne-lapata-2022-zero}, their experiments only cover a few NLs and MRs, impeding comprehensive and unified evaluation on a diverse range of tasks.
Third, due to the lack of a comprehensive CLSP benchmark, the performance of multilingual language models on CLSP is understudied. Some pretrained language models are proposed to solve cross-lingual tasks such as XLM-R~\citep{conneau2019unsupervised} and mT5~\citep{xue2020mt5}, while other large language models are designed for code generation such as Codex~\citep{chen2021evaluating} and BLOOM~\citep{scao2022bloom}. 
However, little research has focused on evaluating models on CLSP.

In this paper, we propose \ours, a unified benchmark for cross-lingual semantic parsing featured with 22 natural languages and 8 meaning representations as summarized in Table~\ref{tab:dataset}. In order to cover a large variety of languages and meaning representations, we first select 9 high-quality CLSP datasets and then clean and format them in a unified manner.
Then, we conduct a comprehensive benchmarking study on three categories of multilingual language models including pretrained encoder-based models augmented with pointer generator (mBERT, XLM-R), pretrained encoder-decoder models (mBART, mT5), and decoder-based large language models (Codex, BLOOM). 
To evaluate these models, we design 6 experiment settings covering various lingual combinations and learning sample scales, including Monolingual (and Monolingual Few-shot), Multilingual, and Cross-lingual Zero-Shot/Few-Shot Transfer.

Our results show that the encoder-decoder model (mT5) yields the best performance on monolingual evaluation compared with other models.
Then, we pick two models with the best monolingual performance (i.e., mT5 and XLM-R) to conduct few-shot and zero-shot cross-lingual transfer learning from English to other low-resource languages. Results show a significant performance gap between monolingual training (Taget NL -> Target NL\footnote{We use A -> B to denote the model finetuned on NL A and tested on NL B.}) and cross-lingual transfer learning (En -> Target NL).
Furthermore, we find that this gap can be significantly reduced by few-shot learning on target NL. We further train these two models in a multilingual setting and find such training can boost the performance in some of the languages, while, however, it usually hurts the performance in English.
Finally, we test two large language models Codex~\citep{chen2021evaluating} and BLOOM~\citep{scao2022bloom}. We find the performance gap of cross-lingual transfer learning is significant for these two models as well.

Our contributions are summarized as follows:
(1) We propose \ours to unify and benchmark 9 datasets covering 5 tasks, 22 natural languages, and 8 meaning representations for cross-lingual semantic parsing; 
(2) We perform a holistic evaluation of 3 groups of state-of-the-art multilingual language models on \ours, demonstrating noticeable performance gaps of cross-lingual transfer models comparing English and other languages; 
(3) We show two effective strategies for boosting performance in low-resource languages: multilingual training and cross-lingual transfer learning.

\begin{table*}[t!]
\centering
\resizebox{\textwidth}{!}{%
\begin{tabular}{@{}lllllllll@{}}
\toprule
Task                  & Dataset    & Meaning Representation        & Language & Executable    & Domain  & Train & Dev & Test \\ 
\midrule
NLI for Databases     & MATIS       & SQL                           & 7        & \tick           & 1       & 4303  & 481    & 444    \\
NLI for Databases     & MGeoQuery   & SQL,Lambda,FunQL,Prolog       & 8        & \tick           & 1       & 548   & 49     & 277     \\
NLI for Databases     & MSpider     & SQL                           & 3        & \tick           & 138     & 8095  & 1034   & --    \\
NLI for Databases     & MNLmaps     & Functional Query Language     & 2        & \tick           & 1       & 1500  & --     & 880   \\
QA on Knowledge Graph & MOvernight  & Lambda Calculus               & 3        & \tick           & 8       & 8754  & 2188   & 2740     \\
QA on Knowledge Graph & MCWQ       & SPARQL               & 4        & \tick           & 1    & 4006 & 733 & 648     \\
QA on Web             & MSchema2QA  & ThingTalk Query Language      & 11       & \tick           & 2       & 8932  &  --   & 971     \\
Task-Oriented DST     & MTOP       & Hierarchical Intent and Slot  & 6        & \cross            & 11      & 5446  & 863 & 1245  \\
Code Generation     & MCoNaLa       & Python  & 4        & \tick            & 1      & 1903  & 476 & 896  \\
\bottomrule
\end{tabular}
}
\caption{Datasets in \ours. We assemble 9 datasets in various domains for 5 semantic parsing tasks. 
It covers 8 meaning representations. The questions cover 22 languages in 15 language families. Train/Dev/Test columns indicate the number of MRs each paired with multiple NLs.
}
\label{tab:dataset}
\end{table*}

\section{\ours Benchmark}
\begin{figure}[t!]
    \centering
    \includegraphics[width=\linewidth]{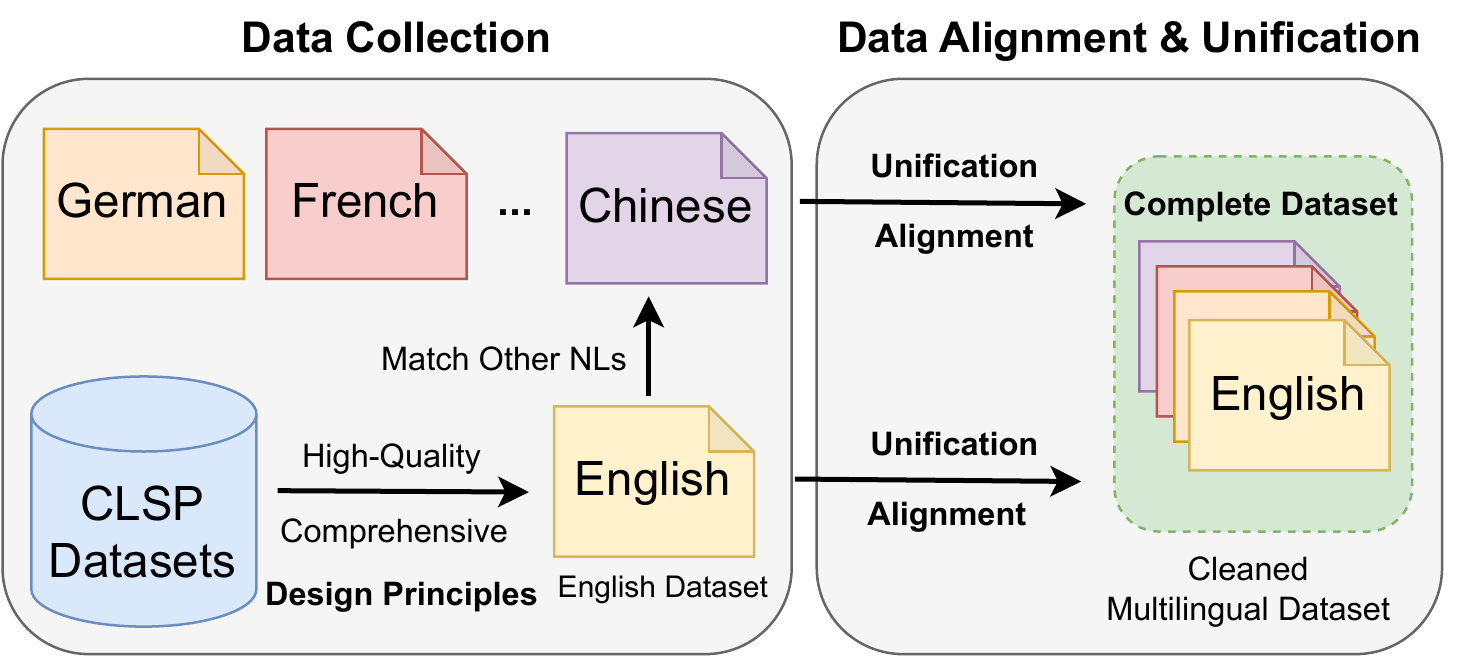}
    \caption{Construction pipeline of \ours. 
    }
    \label{fig:annotation}
\end{figure}

Figure~\ref{fig:annotation} shows the construction pipeline of \ours. We first select 9 CLSP datasets according to our design principles. Then, we collect other NLs of the selected datasets. Finally, we clean the datasets by removing outliers and performing alignment between different languages.

\subsection{Design Principles}
We carefully pick 9 datasets from all available semantic parsing datasets to construct \ours according to two principles. First, the picked datasets need to have \textbf{high quality}, which means they are either annotated by humans or augmented with careful crafting~\citep{moradshahi2020localizing}, and the translation of user inputs are provided by humans instead of machine translation models. Second, \ours needs to be \textbf{comprehensive}~\cite{hu2020xtreme}, which means including diverse NLs and MRs for a broad range of tasks and applications. 

\subsection{Data Collection}
\label{sec:collection}
Table~\ref{tab:dataset} summarizes the characteristics and statistics of different datasets in \ours.

\noindent \textbf{Multilingual ATIS (MATIS)} contains user questions for a flight-booking task. We collect the original English questions from ATIS~\cite{price1990evaluation,dahl1994expanding} and add the translations from~\citet{xu2020end}. For MRs, we focus on the task of Natural Language Interface (NLI) to databases and thus collect SQL from~\citet{iyer2017learning} and \citet{finegan2018improving}.

\noindent \textbf{Multilingual GeoQuery (MGeoQuery)} contains user questions about US geography. We collect original English questions from GeoQuery~\cite{zelle1996learning} and add other translations~\citep{lu2011probabilistic, jones2012semantic, susanto2017semantic}. GeoQuery has several MRs available. We collect Prolog and Lambda Calculus from~\citet{guo2020benchmarking}, FunQL from~\citet{susanto2017semantic}, and SQL from~\citet{finegan2018improving} \footnote{ We report averaged scores of 4 MRs in the tables, unless otherwise specified.}.

\noindent \textbf{Multilingual Spider (MSpider)} is a human-annotated complex and cross-domain text-to-SQL datasets.
We collect Spider~\cite{yu2018spider} with English questions and add other NLs from~\citet{min2019pilot} and~\citet{nguyen2020pilot}.

\noindent \textbf{Multilingual NLmaps (MNLmaps)} is a Natural Language Interface to query the OpenStreetMap database. We collect NLMaps~\cite{lawrence2016nlmaps} in English, and add translations in German~\cite{haas2016corpus}.

\noindent \textbf{Multilingual Overnight (MOvernight)} is a multi-domain semantic parsing dataset in lambda DCS. We include English Overnight~\cite{wang2015building} and add translations from~\citet{sherborne2020bootstrapping}.

\noindent \textbf{Multilingual Schema2QA (MSchema2QA)} is a question answering dataset over schema.org web data in ThingTalk Query Language. We include training examples with all 11 available languages and pair them with the MR in the corresponding language following \citet{moradshahi2020localizing} and \citet{xu2020schema2qa}. To make the dataset size comparable to others, we include 5\% of the training set.

\noindent \textbf{MCWQ} is a multilingual knowledge-based question answering dataset grounded in Wikidata~\cite{cui2021multilingual}. We collect all questions in MCWQ in 4 languages. The split follows maximum compound divergence (MCD)~\citep{DBLP:conf/iclr/KeysersSSBFKMSS20} so that the test set contains novel compounds to evaluate compositionality generalization ability. 

\noindent \textbf{MTOP} is a multilingual semantic parsing dataset for task-oriented dialogs with meaning representations of hierarchical intent and slot annotations~\cite{gupta2018semantic,li2020mtop}. We include examples with all 6 languages and pair the translations with the compositional decoupled representation in the corresponding language.

\noindent \textbf{MCoNaLa} is a multilingual code generation benchmark for Python by extending English CoNaLa~\citep{yin2018learning,wang2022mconala}. We include all 4 languages.

\subsection{Data Alignment and Unification}
We perform data alignment and unification over 9 datasets to construct a unified high-quality benchmark. 
To be specific, for the first 6 datasets introduced in Section~\ref{sec:collection}, because each of them has multiple parts proposed in different work, we merge these parts by aligning the same user question in different languages into the same meaning representation. For the other 3 datasets, we directly use the entire samples since no other parts need to be merged. We also try to unify the language of MRs (e.g., adopting a single form of SQL queries; keeping only one English MR when there is more than one in MTOP). We also remove a few samples in MATIS and MGeoQuery with no MRs. 
We provide more details in Appendix including the examples of each dataset (Table~\ref{tab:example}), data construction (Appendix~\ref{sec:app:construction}), natural languages (Appendix~\ref{sec:app:language}), and meaning representations (Appendix~\ref{sec:app:mr}).

\subsection{Evaluation Metrics}
We evaluate the predicted results using various automatic metrics.
For the Spider dataset, we follow \citet{yu2018spider} and use their proposed tool for evaluation \footnote{All numbers reported in the paper is ``Exact Set Match without Values'' in \url{https://yale-lily.github.io/spider}.}. For the other datasets, we simply use exact matching, i.e., token-by-token string comparison, to see if the prediction is the same as the ground truth label. 
For a fair comparison with state-of-the-art models, we also use the metrics proposed in their models, including Execution Score, Denotation Accuracy, and Code BLEU  (Section~\ref{sec:sota}).

\subsection{Data Analysis}

\paragraph{Natural Languages}
\ours contains diverse and abundant natural languages in both high-resource and low-resource groups, including 22 languages belonging to 15 language families (Appendix~\ref{sec:app:language}).
Most state-of-the-art performances are achieved in English and a few other high-resource languages. However, the lack of information in the low-resource languages brings unanswered questions to model generalization. Therefore, both these 2 types of languages are included in \ours, to form a unified cross-lingual dataset for semantic parsing.
Among these 22 languages, English is the most resourced language with many popular datasets in semantic parsing. Some languages spoken in Western Europe are also relatively high-resource languages, such as German and Spanish. 
We also involve many low-resource languages as well, such as Vietnamese and Thai.  

\paragraph{Meaning Representations}
\ours includes 8 meaning representations for different applications: Prolog, Lambda Calculus, Functional Query Language (FunQL), SQL, ThingTalk Query Language, SPARQL, Python, and Hierarchical intent and slot.
All of them can be executed against underlying databases or knowledge graphs, except for the last one which is designed for complex compositional requests in task-oriented dialogues.
The first four are domain-specific because they contain specific predicates defined for a given domain, while the last four are considered open-domain and open-ontology~\citep{guo2020benchmarking}.
It is also worth noting that these MRs are not equivalent to their general expressiveness.
For example, the ThingTalk query language is a subset of SQL in expressiveness~\cite{moradshahi2020localizing}, and FunQL is less expressive than Lambda Calculus partially due to the lack of variables and quantifiers.

\section{Experiment Setup}
We describe our evaluation settings and models for a comprehensive benchmark study on \ours.
\subsection{Evaluation Settings}
We consider the following 6 settings for training and testing.
\paragraph{Translate-Test.}
We train a model on the English training data and translate target NL test data to English using the public Google NMT system~\cite{wu2016google}. This setting uses one semantic parsing model trained on English but also relies on available machine translation models for other languages. This serves as a strong yet practical baseline for other settings.

\begin{table*}[ht!]
\centering
\resizebox{\textwidth}{!}{
\begin{tabular}{lcccccccccc}
\toprule
    & MATIS & MGeoQuery & MSpider & MNLmaps & MOvernight & MCWQ & MSchema2QA & MTOP & MCoNaLa$^\ddagger$ & Average \\ \midrule
\multicolumn{10}{l}{\textit{Translate-Test}} \\
mT5 & 44.50 & 53.88 & 45.26 & 66.36 & 59.69 & 19.85 & \;\;3.18$^\bigstar$ & 29.78$^\bigstar$ & 8.13 & 36.74 \\
\midrule
\multicolumn{10}{l}{\textit{Monolingual}} \\
mBERT+PTR & 30.63 & 72.18 & 40.40 & 83.82 & 57.47 & 23.46 & 52.53 & 75.41 & \;\;5.87 & 49.09 \\
XLM-R+PTR & 31.31 & 71.41 & 47.30 & 85.17 & 59.10 & 23.53 & 62.37 & 80.36 & \;\;7.69 & 52.03 \\
mBART & 41.93 & 62.29 & 33.31 & 83.19 & 59.60 & 30.02 & 50.35 & 75.76 & \;\;6.78 & 49.25\\
mT5 & \textbf{53.15} & \textbf{74.26} & \textbf{50.73} & \textbf{91.65} & \textbf{66.29} & \textbf{30.15} & \textbf{65.16} & \textbf{81.83} & \textbf{10.29} & \textbf{58.16} \\
\midrule
\multicolumn{10}{l}{\textit{Monolingual Few-Shot}} \\
XLM-R+PTR & 23.44 & 17.91 & 36.04 & 19.77 & 40.74 & \;\;5.64 & \textbf{49.00} & 60.42 & \;\;0.38 & 28.15 \\ 
mT5 & \textbf{24.85} & 25.48 & \textbf{38.10} & 26.93 & \textbf{53.59} & \;\;\textbf{7.68} & 33.27 & \textbf{61.90} & \;\;1.05 & \textbf{30.32} \\
Codex$^\dagger$ & 18.02 & \textbf{31.93} & 30.66 & \textbf{34.26} & \;\;3.43 & \;\;2.93 & 21.62 & 10.08 & \textbf{13.87} & 18.53 \\ 
BLOOM$^\dagger$ & \;\;0.00 & 17.84 & \;\;2.13 & 12.16 & \;\;0.62 & \;\;0.00 & \;\;5.21 & \;\;5.16 & \;\;8.40 & \;\;5.72 \\ 
\midrule
\multicolumn{10}{l}{\textit{Multilingual}} \\
XLM-R+PTR & 39.72 & 71.35 & \textbf{40.20} & 85.91 & 61.03 & \textbf{30.79} & \textbf{61.82} & 81.68 & -- & 59.06 \\
mT5 & \textbf{54.45} & \textbf{76.57} & 32.30 & \textbf{91.31} & \textbf{67.55} & 28.51 & 60.92 & \textbf{82.95} & -- & \textbf{61.82} \\
\midrule
\multicolumn{10}{l}{\textit{Cross-lingual Zero-Shot Transfer}} \\
XLM-R+PTR & \;\;6.05 & \textbf{39.85} & 18.53 & \textbf{60.23} & 36.77 & \;\;\textbf{4.27} & 20.22 & 51.46 & \;\;0.12 & 26.39 \\
mT5 & \textbf{31.85} & 27.35 & \textbf{41.93} & 34.89 & \textbf{52.68} & \;\;4.06 & \textbf{44.04} & \textbf{50.18} & \;\;0.77 & \textbf{31.97} \\
Codex$^\dagger$ & 16.31 & 28.53 & 27.56 & 32.05 & \;\;2.99 & \;\;2.16 & 19.57 & 14.08 & \;\;\textbf{8.35} & 16.84 \\ 
BLOOM$^\dagger$ & \;\;0.00 & 11.29 & \;\;1.70 & \;\;7.05 & \;\;0.38 & \;\;0.00 & \;\;3.93 & \;\;1.67 & \;\;6.16 & \;\;3.58 \\ 
\midrule
\multicolumn{10}{l}{\textit{Cross-lingual Few-Shot Transfer}} \\
XLM-R+PTR & 15.71 & 51.08 & 43.68 & 64.89 & 52.03 & 20.16 & 53.51 & 72.79 & -- & 46.73 \\ 
mT5 & \textbf{49.57} & \textbf{57.31} & \textbf{49.42} & \textbf{71.70} & \textbf{62.53} & \textbf{24.85} & \textbf{59.24} & \textbf{74.83} & -- & \textbf{56.18} \\ 
\bottomrule
\end{tabular}
}
\caption{Results on \ours. We consider 6 settings including 2 Monolingual, 1 Multilingual, and 2 Cross-lingual settings, and one Translate-Test setting. Each number is averaged across different languages in that dataset. $^\dagger$ Codex/BLOOM are evaluated on only two settings as we apply 8-shot in-context learning without finetuning the model parameters. $^\ddagger$ Two settings are not applicable to MCoNaLa because it has no training set on NLs other than English. $^\bigstar$ Translate-Test performances on MSchem2QA and MTOP are especially low because the MR of these data also contains tokens in target languages.}
\label{tab:experiment}
\end{table*}

\paragraph{Monolingual.} 
We train a monolingual model on each target NL training data. This setting creates one model per target NL. In addition to benchmarking them, we design this setting for two reasons: (1) It helps the comparison between monolingual and cross-lingual performance; (2) We pick the best models from this setting to further conduct cross-lingual and few-shot/zero-shot experiments.
Additionally, since some target NL training data can be expensive to obtain, we also test a \textbf{Monolingual Few-shot} setting by training monolingual models with only 10\% training data. 

\paragraph{Multilingual.}
Thanks to the progress in multilingual embeddings and pretrained multilingual language models, we can train one multilingual model on all NL training data. This setting uses only one model to serve all NLs.

\paragraph{Cross-lingual Zero-shot Transfer.}
Models are trained only on English NL data and then tested on a target-NL test set. This setting uses one model for all target NLs and evaluates the cross-lingual transfer ability without any target-NL training data.
Besides, to test the value of additional target NL training data, we finetune the model on 10\% target-NL training data. This \textbf{Cross-lingual Few-shot Transfer} setting creates one model per target NL. We use these two settings to evaluate the capability of the model to transfer from a fine-tuned model of high-resource NL to a low-resource test set.

\subsection{Models}
\label{sec:models}
We evaluate three different groups of multilingual language models on \ours.

\paragraph{Multilingual Pretrained Encoders with Pointer-based Decoders (Enc-PTR).}
The first group is multilingual pretrained encoders with decoders augmented with pointers.
Both encoders and decoders use Transformers~\cite{vaswani2017attention}.
The decoder uses pointers to copy entities from natural language inputs to generate meaning representations~\cite{rongali2020don,prakash2020compressing}.
We use two types of multilingual pretrained encoders, mBERT~\cite{devlin2018bert} and XLM-R~\cite{conneau2019unsupervised}, and both are trained on web data covering over 100 languages.

\paragraph{Multilingual Pretrained Encoder-Decoder Models (Enc-Dec).}
The second group uses pretrained encoder-decoder models, including mBART~\cite{chipman2022mbart} and mT5~\cite{xue2020mt5} which uses text-to-text denoising objective for pretraining over multilingual corpora.

\paragraph{Multilingual Large Language Models (LLMs).} 
The third group is multilingual large language models based on GPT~\citep{brown2020language} including Codex~\citep{chen2021evaluating} and BLOOM~\citep{scao2022bloom}. Codex is fine-tuned on publicly available code from GitHub. While it is not trained on a multilingual corpus, it has shown cross-lingual semantic parsing capabilities~\cite{shi2022xricl}. BLOOM is a 176B-parameter multilingual language model pretrained on 46 natural and 13 programming languages from the ROOTS corpus~\citep{laurenccon2022bigscience}. We mainly use these models to evaluate the ability of few-shot learning using in-context learning without any further finetuning. Specifically, we append 8 samples and the test query to predict the MR. For Monolingual Few-shot, samples and the query are in the same NL, while for Cross-lingual Zero-shot Transfer, samples are in English and the query is in the target NL.

\section{Results and Analysis}
\label{sec:overall}
Table~\ref{tab:experiment} shows the performance of all 6 models on 6 settings. 
Our results and analysis aim to answer the following research questions:
\begin{itemize}[noitemsep,topsep=0pt,parsep=0pt,partopsep=0pt]
    \item RQ 1: What is the best model and training strategy for performance, and how does it compare with previous state-of-the-art? (Section~\ref{sec:anamono},~\ref{sec:sota})
    \item RQ 2: How capable are the current multilingual LLMs on the task of CLSP? (Section~\ref{sec:anacodex})
    \item RQ 3: What is the effect of few-shot learning? (Section~\ref{sec:anafew})
    \item RQ 4: What is the effect of multilingual learning? (Section~\ref{sec:anamulti})
    \item RQ 5: What is the effect of cross-lingual transfer learning? (Section~\ref{sec:anacross})
    \item RQ 6: How performance varies across different natural languages and meaning representations? (Section~\ref{ananls},~\ref{sec:anamrs})
\end{itemize}

\subsection{Analysis of Monolingual}
\label{sec:anamono}
 We obtain the following main findings on Monolingual setting:

Enc-Dec (mT5) obtains the best performance. Among the two transformer-based pointer generators, XLM-R+Transformer (XLM-R+PTR) (52.03\footnote{If not specified, the numbers in this section are the averaged exact matching scores across all NLs.}) performs slightly better than mBERT+Transformer (mBERT+PTR) (49.09). Among mBART and mT5, mT5 (58.16) outperforms mBART (49.25) by a large margin. Besides, although mT5 outperforms XLM-R by 6.13, XLM-R is still able to outperform mBART by 2.78. Thus, we pick mT5 among mT5/mBART, and XLM-R among XLM-R/mBERT to conduct the experiments on the other settings.

Next, we evaluate mT5 model on Translation-Test setting. As shown in the table, mT5 in Monolingual setting outperforms Translation-Test by a large margin (58.16 vs. 36.74). This shows that multilingual language models are more effective than Translation-Test methods. In other words, it is necessary to train a multilingual model even though we have a high-quality translation system. 

\subsection{Comparison with SOTA}
\label{sec:sota}
Table~\ref{tab:sota} lists the performance of mT5 in Monolingual setting with the previous state-of-the-art. Some of the previous work use denotation accuracy and execution accuracy which are different from the exact match we use. To make our results comparable with previous work, we apply the evaluation tools of previous work to \ours. 
As shown in the table, Enc-Dec (mT5) outperforms previous work on all NLs of MSchema2QA, MCWQ, MNLMaps, MATIS datasets and obtains comparable results on the others.

\begin{table*}[t!]
\centering
\resizebox{\textwidth}{!}{
\begin{tabular}{@{}ccllc@{}}
\toprule
\multicolumn{1}{c}{Dataset}  & Language & \multicolumn{1}{c}{SOTA~(Source)} & \multicolumn{1}{l}{\ours} & Metric              \\ \midrule
\multirow{6}{*}{MSpider}     & English       & 77.10~\citep{li2023graphix}                   & 67.60                                    & Exact Match         \\
                             & English       & 81.00~\citep{li2023graphix}                    & 69.10                                    & Execution           \\
                             & Vietnamese       & 69.00~\citep{shi-etal-2022-cross}                    & 43.00                     & Exact Match         \\
                             & Vietnamese       & 64.50~\citep{shi-etal-2022-cross}                    & 42.00                     & Execution           \\
                             & Chinese       & 66.1$^\bigstar$~\citep{shi-etal-2022-cross}                    & 39.90                                    & Exact Match         \\
                             
                             \midrule
\multirow{10}{*}{MSchema2QA} & Arabic       & 29.17~\citep{moradshahi2020localizing}                    & 53.55                                    & Exact Match         \\
                             & German       & 51.84~\citep{moradshahi2020localizing}                    & 72.19                                    & Exact Match         \\
                             & Spanish       & 56.01~\citep{moradshahi2020localizing}                    & 68.69                                    & Exact Match         \\
                             & Farsi       & 54.88~\citep{moradshahi2020localizing}                    & 60.25                                    & Exact Match         \\
                             & Finnish       & 52.43~\citep{moradshahi2020localizing}                    & 68.28                                    & Exact Match         \\
                             & Italian       & 54.87~\citep{moradshahi2020localizing}                    & 67.97                                    & Exact Match         \\
                             & Japanese       & 46.27~\citep{moradshahi2020localizing}                    & 62.41                                    & Exact Match         \\
                             & Polish       & 49.69~\citep{moradshahi2020localizing}                    & 60.87                                    & Exact Match         \\
                             & Turkish       & 56.84~\citep{moradshahi2020localizing}                    & 70.03                                    & Exact Match         \\
                             & Chinese       & 36.60~\citep{moradshahi2020localizing}                    & 56.54                                    & Exact Match         \\\midrule
\multirow{4}{*}{MCWQ}        & English       & 27.70~\citep{cui2022compositional}                    & 39.29                                    & Exact Match         \\
                             & Hebrew       & 16.60~\citep{cui2022compositional}                    & 33.02                                    & Exact Match         \\
                             & Kannada       & 16.60~\citep{cui2022compositional}                    & 23.74                                    & Exact Match         \\
                             & Chinese       & 23.00~\citep{cui2022compositional}                    & 24.56                                    & Exact Match         \\\midrule
\multirow{2}{*}{MNLMaps}     & English       & 85.70~\citep{duong2017multilingual}                    & 92.73                                    & Exact Match         \\
                             & German       & 83.00~\citep{duong2017multilingual}                    & 90.57                                    & Exact Match         \\\midrule
\multirow{6}{*}{MATIS}       & English       & 77.20~\citep{sherborne2023meta}                    & 83.78                                    & Denotation accuracy \\
                             & Farsi       & 67.80~\citep{sherborne2023meta}                    & 80.59                                    & Denotation accuracy \\
                             & Portuguese       & 66.10~\citep{sherborne2023meta}                    & 78.60                                    & Denotation accuracy \\
                             & Spanish       & 64.10~\citep{sherborne2023meta}                    & 76.58                                    & Denotation accuracy \\
                             & German       & 66.60~\citep{sherborne2023meta}                    & 80.63                                    & Denotation accuracy \\
                             & Chinese       & 64.90~\citep{sherborne2023meta}                    & 78.38                                    & Denotation accuracy \\\midrule
\multirow{8}{*}{MGeoQuery$^\dagger$}   & English       & 90.00~\citep{zou2018learning}                    & 79.06                     & Denotation accuracy \\
                             & Thai       & 86.10~\citep{zou2018learning}                    & 72.56                     & Denotation accuracy \\
                             & German       & 76.80~\citep{zou2018learning}                    & 73.29                     & Denotation accuracy \\
                             & Greek       & 83.20~\citep{zou2018learning}                    & 76.90                     & Denotation accuracy \\
                             & Chinese       & 82.10~\citep{zou2018learning}                    & 75.81                     & Denotation accuracy \\
                             & Indonesian       & 83.90~\citep{zou2018learning}                    & 80.14                     & Denotation accuracy \\
                             & Swedish       & 83.90~\citep{zou2018learning}                    & 79.78                     & Denotation accuracy \\
                             & Farsi       & 76.80~\citep{zou2018learning}                    & 69.68                     & Denotation accuracy \\\midrule
\multirow{3}{*}{MOvernight}  & English       & 81.90~\citep{sherborne2021zero}                    & 69.38$^\ddagger$                     & Denotation accuracy \\
                             & German       & 66.20~\citep{sherborne2021zero}                    & 66.90$^\ddagger$                     & Denotation accuracy \\
                             & Chinese       & 66.00~\citep{sherborne2021zero}                    & 62.59$^\ddagger$                     & Denotation accuracy \\\midrule
\multirow{3}{*}{MCoNaLa}     & Russian       & 9.56~\citep{wang2022mconala}                     & 6.38                                     & Code BLEU-4         \\
                             & Spanish       & 2.64~\citep{wang2022mconala}                     & 2.55                                     & Code BLEU-4         \\
                             & Japanese       & 9.90~\citep{wang2022mconala}                     & 7.66                                     & Code BLEU-4         \\ \bottomrule 
\end{tabular}
}
\caption{Comparison between mT5 monolingual and state-of-the-art models, except that MCoNaLa dataset uses cross-lingual zero-shot settings because the dataset only contains English training samples. mT5 obtains better or comparable performance on all datasets. $^\bigstar$ Previous SOTA model only contains exact match scores for Chinese. $^\dagger$ The SOTA model of MGeoQuery uses Lambda as MR while \ours uses SQL. $^\ddagger$ The SOTA model of MOvernight uses denotation accuracy and \ours uses exact match. }
\label{tab:sota}
\end{table*}

\subsection{Analysis of Codex and BLOOM}
\label{sec:anacodex}
We evaluate Codex and BLOOM to test the performance of in-context learning of large language models. As shown in Table~\ref{tab:experiment}, LLMs (Codex and BLOOM) are outperformed by mT5 model by a large margin for both Few-shot (11.79/24.60) and Zero-shot (15.13/28.39) settings. This suggests that multilingual LLMs are still inadequate for cross-lingual semantic parsing tasks.

\subsection{Comparison between Few-shot Settings}
\label{sec:anafew}
We also test the Enc-Dec (mT5) and Enc-PTR (XLM-R) models on two types of few-shot experiments, including Monolingual and Cross-lingual Few-Shot.

As can be seen, mT5 of cross-lingual few-shot outperforms monolingual few-shot by a large 22.21 exact match score (excluding MCoNaLa), while XLM-R has a smaller gain of 15.12. We can summarize two observations: 1) pretraining on the English NL can significantly boost the performance of few-shot on target NLs (En + Target Few-shot -> Target NL), and 2) the model with higher cross-lingual capability gains more improvement, such as mT5 gains more than XLM-R. 
Both observations demonstrate the capability of cross-lingual models to transfer knowledge from the source to the target NLs. 

\subsection{Analysis of Multilingual Training}
\label{sec:anamulti}
\begin{figure}[t!]
    \centering
    \includegraphics[width=\linewidth]{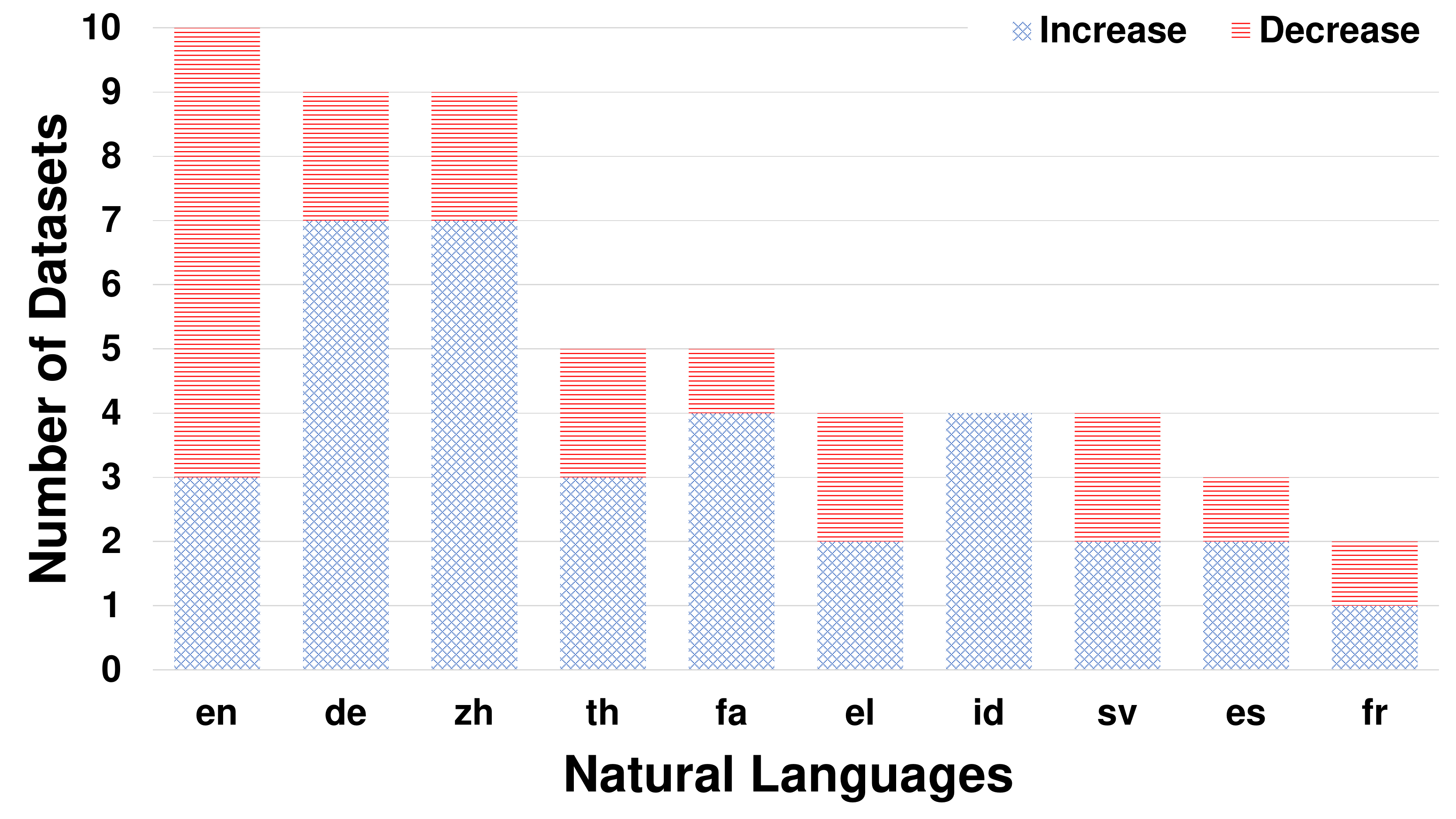}
    \caption{Effect of multilingual training with mT5 on different NLs. X-axis is the NL that was included in at least two datasets. Y-axis is the number of datasets that the performance increases/decreases of this NL after multilingual training. Performance of English (high resource NLs) are easier to drop in multilingual training.
    }
    \label{fig:lang}
\end{figure}
We compare the performance of Monolingual and Multilingual settings. As can be seen in Table~\ref{tab:experiment}, mT5 improves by 2.31 on MGeoQuery, and XLM-R improves by 8.41 on MATIS dataset. This demonstrates that Enc-Dec/Enc-PTR (mT5/XLM-R) can be improved by training in a mixture of various languages. However, not all datasets can boost performance via such training. The average change of mT5/XLM-R is around -2/+2 points. 

We further explore the reason for the performance drop in multilingual training. As shown in Figure~\ref{fig:lang}, most of the major NLs can obtain performance gain, except that English performance drops in 7 datasets and gains in 3 datasets. This is known as "Curse of Multilinguality"~\citep{pfeiffer-etal-2022-lifting}. Similarly in CLSP, performance of English (high resource NLs) is easier to drop in multilingual training.

\subsection{Cross-lingual Performance Gap}
\label{sec:anacross}

\begin{figure}[t!]
    \centering
    \small
    \includegraphics[width=\linewidth]{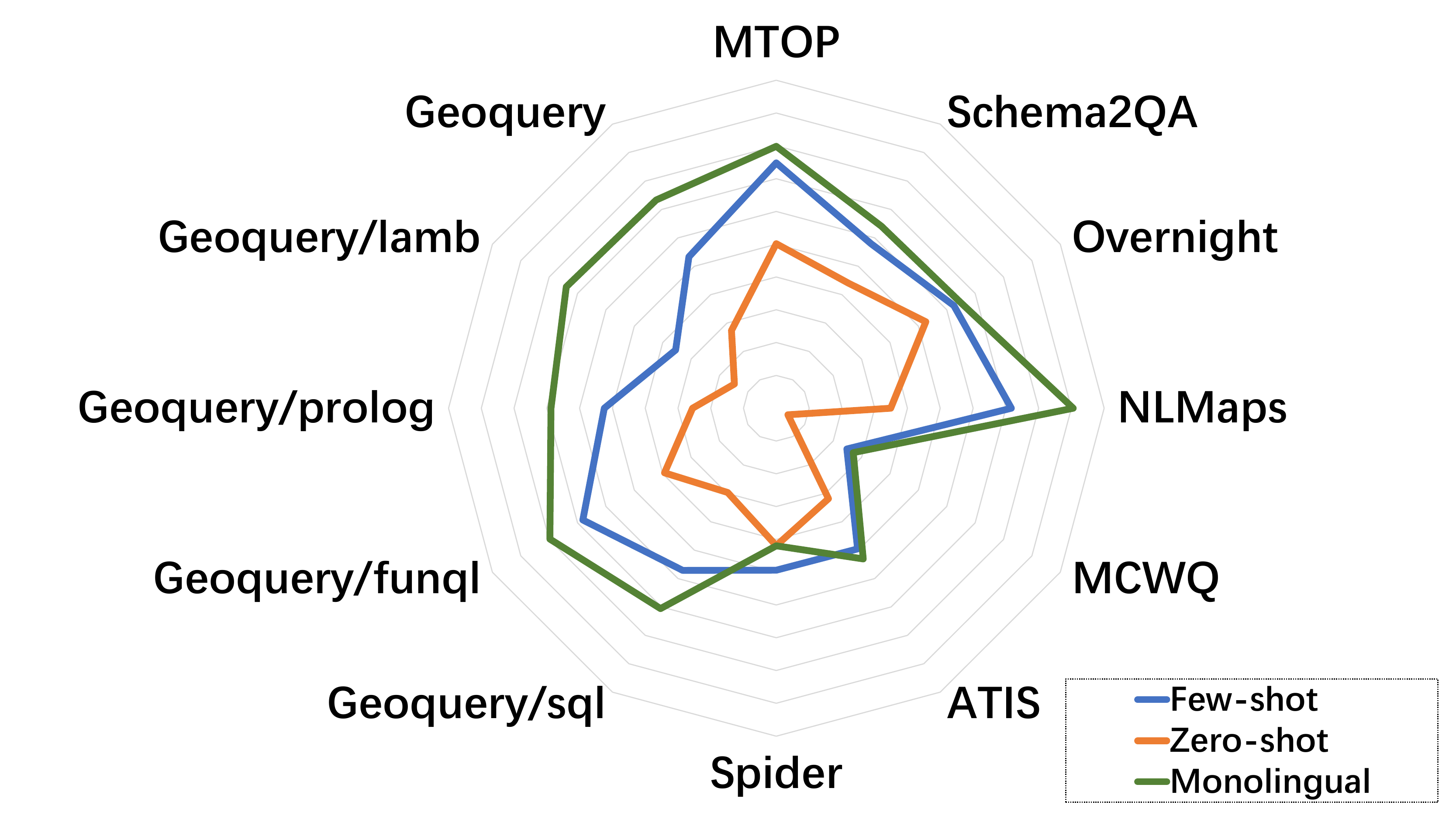}
    \caption{The performance of cross-lingual Few/Zero-shot (mT5) on different datasets and languages. MGeoQuery/* indicates a single MR; MGeoQuery is the averaged score across 4 MRs. Each neighbor grey circle has a 10 score difference, and the center of the circle indicates a 0 score. The cross-lingual transfer performance gap is significant for the zero-shot setting. However, few-shot training shrinks this gap greatly. 
    }
    \label{fig:gap}
\end{figure}

To examine the transfer ability of the cross-lingual models, we investigate the performance difference between the Monolingual and Cross-lingual Few/Zero-shot for each dataset using mT5. As shown in Figure~\ref{fig:gap}, 
by examining the distance between green and orange lines, we find that for the zero-shot setting, the cross-lingual transfer performance gap is significant, which is even larger than 50\% on the NLmaps dataset, demonstrating the limitation of current cross-lingual models.
However, by examining the difference between orange and blue lines, we also find that using even 10\% of samples in target data, the transfer gap will be shortened rapidly. The few-shot gap usually shrinks to around half of the zero-shot gap, e.g., the Schema2QA dataset. For MATIS, the gap even shrinks to around 5 which is very close to the performance of the monolingual setting.

\subsection{Analysis over Natural Languages}
\label{ananls}
\begin{figure}[t!]
    \centering
    \includegraphics[width=\linewidth]{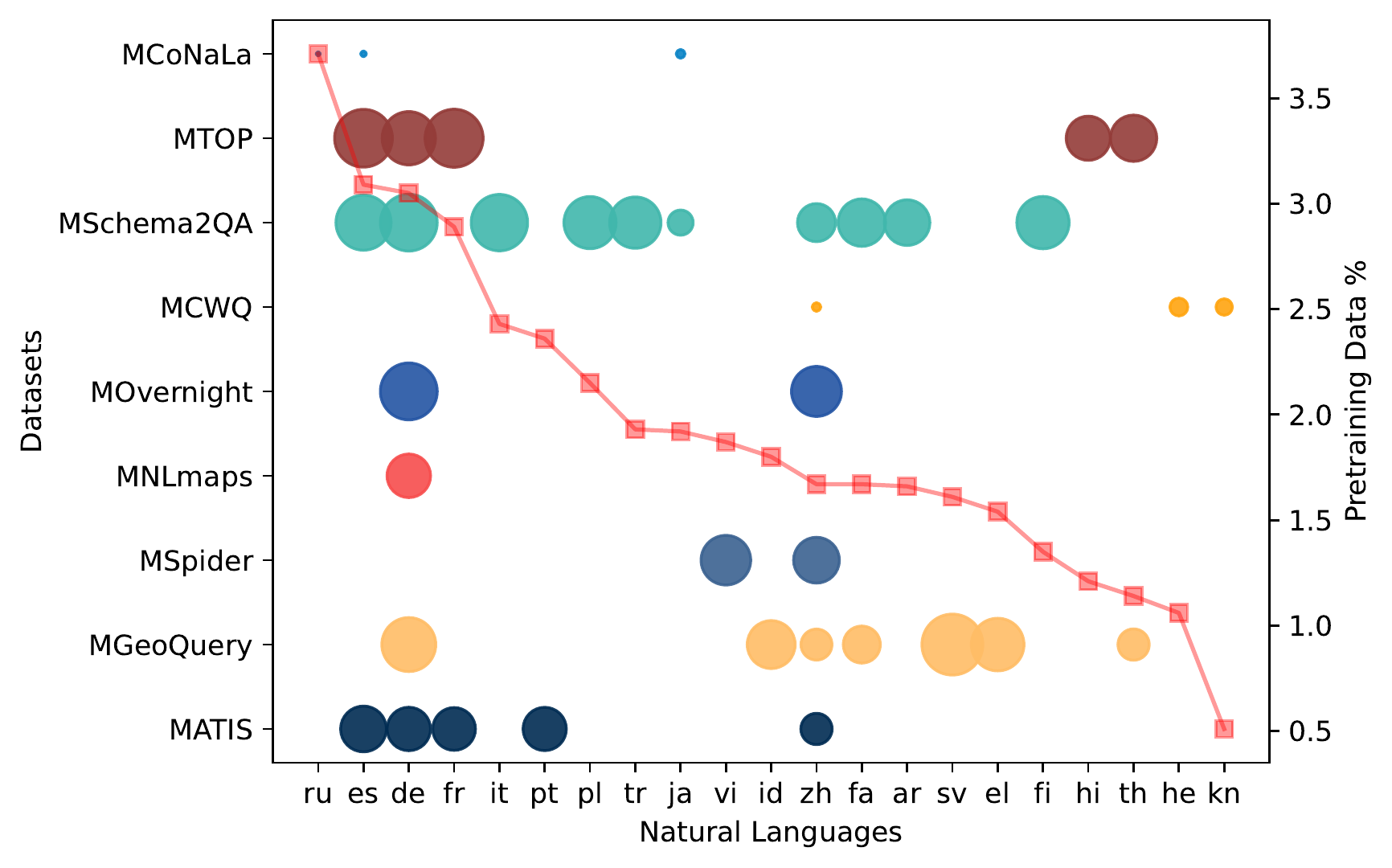}
    \caption{Left vertical axis: The performance of cross-lingual zero-shot mT5 models on different datasets over different languages. Larger dots indicate higher accuracy. Right vertical axis: Red line indicates the percentage of different languages in the mT5 training data. Chinese/German has the largest/smallest performance loss for transfer learning. Additionally, performance and pretraining data size have no evident correlation.} 
    \label{fig:cross}
\end{figure}

We pick the best model mT5 and analyze its performance in the zero-shot setting in Figure~\ref{fig:cross}. Results show that the performance  of Chinese transfer learning (En -> Zh) and English monolingual training (En -> En) usually is the largest compared with transfer learning of other NLs.
On the other hand, German usually has the smallest transfer performance loss. This is probably because of two reasons. First, the Chinese data source is less than German when pretraining on mT5. Second, the language family of English is closer to German (IE: Germanic) compared with Chinese (Sino-Tibetan). This phenomenon is discussed in \citet{hu2020xtreme}, and we find this conclusion also holds for cross-lingual semantic parsing tasks.

\subsection{Analysis over Meaning Representations}
\label{sec:anamrs}
Table~\ref{tab:mr} shows the performance of mT5 on various MRs in MGeoQuery. In almost all languages, FunQL outperforms the other three meaning representations, and SQL obtains the worst performance. This is consistent with the observation of \citet{guo2020benchmarking}.
We speculate that there are two possible reasons: (1) the grammar of SQL is more complex than the others, and FunQL enjoys much easier grammar~\citep{li-etal-2022-exploring-secrets}, and (2) FunQL contains a number of brackets that provide information of structure to the models~\citep{shu-etal-2021-logic}.

\section{Related Work}
\paragraph{Cross-lingual Semantic Parsing}
Most semantic parsing datasets are originally in English such as GeoQuery~\cite{zelle1996learning}, ATIS~\cite{finegan2018improving}, Overnight~\cite{wang2015building}, and Spider~\cite{yu2018spider}.
Cross-lingual Semantic Parsing datasets are usually constructed by translating the English user questions into other languages~\cite{dou2022multispider,athiwaratkun2022multi}.
For example, \citet{lu2011probabilistic} translate GeoQuery English queries to create a Chinese version. \citet{min2019pilot} and \citet{nguyen2020pilot} create Chinese and the Vietnamese translation of Spider.
However, existing CLSP datasets follow different formats and are independently studied as separate efforts. We aim to provide a unified benchmark and modeling framework to facilitate systematic evaluation and generalizable methodology.
\begin{table}[t!]
\begin{tabular}{@{}lcccc@{}}
\toprule
           & \multicolumn{1}{l}{SQL}  & \multicolumn{1}{l}{Prolog} & \multicolumn{1}{l}{Lambda} & \multicolumn{1}{l}{FunQL}\\ \midrule
English    & 76.50                             & 81.59                      & 76.50             & \textbf{89.89}                    \\
German     & 68.23                    & 64.26                      & \textbf{72.20}             & 71.83                             \\
Thai       & 68.59                             & 63.90                      & 70.04             & \textbf{76.17}                    \\
Chinese    & 70.04                             & 63.18                      & 74.37             & \textbf{77.62}                    \\
Farsi      & 64.98                             & 61.73                      & 64.62             & \textbf{75.45}                    \\
Greek      & 71.84                             & 75.81                      & 78.70             & \textbf{85.92}                    \\
Indonesian & 75.09                             & 75.09                      & 78.34             & \textbf{87.00 }                   \\
Swedish    & 75.45                             & 77.26                      & 79.78             & \textbf{84.48}                    \\
Average        & 71.34                             & 70.35                      & 74.32             & \textbf{81.04}                    \\ \bottomrule
\end{tabular}
\caption{Monolingual performance of mT5 on MGeoQuery. FunQL/SQL obtains the best/worst performance.
}
\label{tab:mr}
\vspace{-3mm}
\end{table}

\paragraph{Multilingual Language Models}
There has been significant progress in multilingual language models.
MUSE~\cite{conneau2017word} aligns monolingual word embeddings in an unsupervised way without using any parallel corpora.
XLM~\cite{lample2019cross} is a pretrained language model based on RoBERTa~\cite{liu2019roberta} which offers cross-lingual contextualized word representations.
Similarly, mBERT is developed as the multilingual version of BERT~\citet{devlin2018bert}.
XLM-R~\cite{conneau2019unsupervised} outperforms mBERT and XLM in sequence labeling, classification, and question answering.
Focusing on sequence-to-sequence tasks such as machine translation, mBART~\cite{liu2020multilingual} extends BART by introducing multilingual denoising pretraining.
mT5~\cite{xue2020mt5} extends T5 by pretraining on the multilingual dataset mC4. Multilingual large language models have been proposed such as BLOOM~\cite{scao2022bloom} and XGLM~\cite{lin2022few}. From multilingual embeddings to multilingual large language models, there have been more effective representations as well as more languages covered~\cite{srivastava2022beyond}. 
We aim to systematically evaluate these models on CLSP, which is understudied by existing work.

\paragraph{Cross-lingual NLP Benchmarks}
Cross-lingual benchmarks have been established in many NLP tasks. 
XNLI is a large-scale corpus aimed to provide a standardized evaluation set \cite{conneau2018xnli}. \citet{hu2020xtreme} developed XTREME to evaluate how well multilingual representations in 40 languages can generalize. XGLUE is another dataset used to implement evaluation in various cross-lingual tasks \cite{liang2020xglue}. MLQA \cite{lewis2019mlqa}, XQuAD \cite{artetxe2019cross}, and XOR QA \cite{asai2020xor} are three evaluation frameworks for cross-lingual question answering. \citet{sun2020clirmatrix} introduce CLIRMatrix by collecting multilingual datasets from Wikipedia for cross-lingual information retrieval~\cite{zbib2019neural,oard2019surprise,zhang2019improving,shi2021cross,chen2021cross}.
For cross-lingual summarization, NLCS was built by \citet{zhu2019ncls} to tackle the problem of the divided summarization and translation.
Nonetheless, there is no unified benchmark for CLSP, and thus we are unable to calibrate the performance of multilingual language models on CLSP.

\section{Conclusion}
We build \ours, a unified benchmark for cross-lingual semantic parsing with multiple natural languages and meaning representations. We conduct a comprehensive benchmark study on three representative types of multilingual language models. Our results show that mT5 with monolingual training yields the best performance, while notably multilingual LLMs are still inadequate to perform cross-lingual semantic parsing tasks. Moreover, the performance gap between monolingual training and cross-lingual transfer learning is still significant. These findings call for both improved semantic parsing capabilities of multilingual LLMs and stronger cross-lingual transfer learning techniques for semantic parsing.

\section*{Limitations}
While we cover a wide range of different factors of cross-lingual semantic parsing (e.g., tasks, datasets, natural languages, meaning representations, domains), we cannot include all possible dimensions along with these aspects.
Furthermore, we focus on the linguistic generalization ability for semantic parsing because the questions are translated from the English datasets.
In the future, we will explore questions raised by native speakers in each language to study the model ability under variations in cultural backgrounds and information-seeking needs.

\section*{Acknowledgment}
We thank Victoria Lin, Bailin Wang, Robin Jia, Ice Pasupat, Tianze Shi, Bing Xiang, Luke Zettlemoyer for their early feedback and discussions. We thank Peng Shi, Yucheng Nie, Junru Liu, Tom Sherborne, Harsh Maniar, Xiangyu Dong, Chen Wang, Songlin Hou, Haoran Zhang, Nan Zhang, and Sarkar Das for their valuable help and comments. 
\bibliography{custom}

\begin{thebibliography}{76}
\expandafter\ifx\csname natexlab\endcsname\relax\def\natexlab#1{#1}\fi

\bibitem[{Artetxe et~al.(2019)Artetxe, Ruder, and Yogatama}]{artetxe2019cross}
Mikel Artetxe, Sebastian Ruder, and Dani Yogatama. 2019.
\newblock On the cross-lingual transferability of monolingual representations.
\newblock \emph{arXiv preprint arXiv:1910.11856}.

\bibitem[{Asai et~al.(2020)Asai, Kasai, Clark, Lee, Choi, and
  Hajishirzi}]{asai2020xor}
Akari Asai, Jungo Kasai, Jonathan~H Clark, Kenton Lee, Eunsol Choi, and
  Hannaneh Hajishirzi. 2020.
\newblock Xor qa: Cross-lingual open-retrieval question answering.
\newblock \emph{arXiv preprint arXiv:2010.11856}.

\bibitem[{Athiwaratkun et~al.(2022)Athiwaratkun, Gouda, Wang, Li, Tian, Tan,
  Ahmad, Wang, Sun, Shang et~al.}]{athiwaratkun2022multi}
Ben Athiwaratkun, Sanjay~Krishna Gouda, Zijian Wang, Xiaopeng Li, Yuchen Tian,
  Ming Tan, Wasi~Uddin Ahmad, Shiqi Wang, Qing Sun, Mingyue Shang, et~al. 2022.
\newblock Multi-lingual evaluation of code generation models.
\newblock \emph{arXiv preprint arXiv:2210.14868}.

\bibitem[{Berant et~al.(2013)Berant, Chou, Frostig, and
  Liang}]{berant2013semantic}
Jonathan Berant, Andrew Chou, Roy Frostig, and Percy Liang. 2013.
\newblock Semantic parsing on freebase from question-answer pairs.
\newblock In \emph{Proceedings of the 2013 conference on empirical methods in
  natural language processing}, pages 1533--1544.

\bibitem[{Brown et~al.(2020)Brown, Mann, Ryder, Subbiah, Kaplan, Dhariwal,
  Neelakantan, Shyam, Sastry, Askell et~al.}]{brown2020language}
Tom Brown, Benjamin Mann, Nick Ryder, Melanie Subbiah, Jared~D Kaplan, Prafulla
  Dhariwal, Arvind Neelakantan, Pranav Shyam, Girish Sastry, Amanda Askell,
  et~al. 2020.
\newblock Language models are few-shot learners.
\newblock \emph{Advances in neural information processing systems},
  33:1877--1901.

\bibitem[{Chen et~al.(2021{\natexlab{a}})Chen, Tworek, Jun, Yuan, Pinto,
  Kaplan, Edwards, Burda, Joseph, Brockman et~al.}]{chen2021evaluating}
Mark Chen, Jerry Tworek, Heewoo Jun, Qiming Yuan, Henrique Ponde de~Oliveira
  Pinto, Jared Kaplan, Harri Edwards, Yuri Burda, Nicholas Joseph, Greg
  Brockman, et~al. 2021{\natexlab{a}}.
\newblock Evaluating large language models trained on code.
\newblock \emph{arXiv preprint arXiv:2107.03374}.

\bibitem[{Chen et~al.(2021{\natexlab{b}})Chen, Kedzie, Nair,
  Galu{\v{s}}{\v{c}}{\'a}kov{\'a}, Zhang, Oard, and McKeown}]{chen2021cross}
Yanda Chen, Chris Kedzie, Suraj Nair, Petra Galu{\v{s}}{\v{c}}{\'a}kov{\'a},
  Rui Zhang, Douglas~W Oard, and Kathleen McKeown. 2021{\natexlab{b}}.
\newblock Cross-language sentence selection via data augmentation and rationale
  training.
\newblock \emph{arXiv preprint arXiv:2106.02293}.

\bibitem[{Chipman et~al.(2022)Chipman, George, McCulloch, and
  Shively}]{chipman2022mbart}
Hugh~A Chipman, Edward~I George, Robert~E McCulloch, and Thomas~S Shively.
  2022.
\newblock mbart: Multidimensional monotone bart.
\newblock \emph{Bayesian Analysis}, 17(2):515--544.

\bibitem[{Conneau et~al.(2019)Conneau, Khandelwal, Goyal, Chaudhary, Wenzek,
  Guzm{\'a}n, Grave, Ott, Zettlemoyer, and Stoyanov}]{conneau2019unsupervised}
Alexis Conneau, Kartikay Khandelwal, Naman Goyal, Vishrav Chaudhary, Guillaume
  Wenzek, Francisco Guzm{\'a}n, Edouard Grave, Myle Ott, Luke Zettlemoyer, and
  Veselin Stoyanov. 2019.
\newblock Unsupervised cross-lingual representation learning at scale.
\newblock \emph{arXiv preprint arXiv:1911.02116}.

\bibitem[{Conneau et~al.(2017)Conneau, Lample, Ranzato, Denoyer, and
  J{\'e}gou}]{conneau2017word}
Alexis Conneau, Guillaume Lample, Marc'Aurelio Ranzato, Ludovic Denoyer, and
  Herv{\'e} J{\'e}gou. 2017.
\newblock Word translation without parallel data.
\newblock \emph{arXiv preprint arXiv:1710.04087}.

\bibitem[{Conneau et~al.(2018)Conneau, Lample, Rinott, Williams, Bowman,
  Schwenk, and Stoyanov}]{conneau2018xnli}
Alexis Conneau, Guillaume Lample, Ruty Rinott, Adina Williams, Samuel~R Bowman,
  Holger Schwenk, and Veselin Stoyanov. 2018.
\newblock Xnli: Evaluating cross-lingual sentence representations.
\newblock \emph{arXiv preprint arXiv:1809.05053}.

\bibitem[{Cui et~al.(2021)Cui, Aralikatte, Lent, and
  Hershcovich}]{cui2021multilingual}
Ruixiang Cui, Rahul Aralikatte, Heather Lent, and Daniel Hershcovich. 2021.
\newblock Multilingual compositional wikidata questions.
\newblock \emph{arXiv preprint arXiv:2108.03509}.

\bibitem[{Cui et~al.(2022)Cui, Aralikatte, Lent, and
  Hershcovich}]{cui2022compositional}
Ruixiang Cui, Rahul Aralikatte, Heather Lent, and Daniel Hershcovich. 2022.
\newblock Compositional generalization in multilingual semantic parsing over
  wikidata.
\newblock \emph{Transactions of the Association for Computational Linguistics},
  10:937--955.

\bibitem[{Dahl et~al.(1994)Dahl, Bates, Brown, Fisher, Hunicke-Smith, Pallett,
  Pao, Rudnicky, and Shriberg}]{dahl1994expanding}
Deborah~A Dahl, Madeleine Bates, Michael~K Brown, William~M Fisher, Kate
  Hunicke-Smith, David~S Pallett, Christine Pao, Alexander Rudnicky, and
  Elizabeth Shriberg. 1994.
\newblock Expanding the scope of the atis task: The atis-3 corpus.
\newblock In \emph{HUMAN LANGUAGE TECHNOLOGY: Proceedings of a Workshop held at
  Plainsboro, New Jersey, March 8-11, 1994}.

\bibitem[{Devlin et~al.(2018)Devlin, Chang, Lee, and
  Toutanova}]{devlin2018bert}
Jacob Devlin, Ming-Wei Chang, Kenton Lee, and Kristina Toutanova. 2018.
\newblock Bert: Pre-training of deep bidirectional transformers for language
  understanding.
\newblock \emph{arXiv preprint arXiv:1810.04805}.

\bibitem[{Dou et~al.(2022)Dou, Gao, Pan, Wang, Che, Zhan, and
  Lou}]{dou2022multispider}
Longxu Dou, Yan Gao, Mingyang Pan, Dingzirui Wang, Wanxiang Che, Dechen Zhan,
  and Jian-Guang Lou. 2022.
\newblock Multispider: Towards benchmarking multilingual text-to-sql semantic
  parsing.
\newblock \emph{arXiv preprint arXiv:2212.13492}.

\bibitem[{Duong et~al.(2017)Duong, Afshar, Estival, Pink, Cohen, and
  Johnson}]{duong2017multilingual}
Long Duong, Hadi Afshar, Dominique Estival, Glen Pink, Philip~R Cohen, and Mark
  Johnson. 2017.
\newblock Multilingual semantic parsing and code-switching.
\newblock In \emph{Proceedings of the 21st Conference on Computational Natural
  Language Learning (CoNLL 2017)}, pages 379--389.

\bibitem[{Finegan-Dollak et~al.(2018)Finegan-Dollak, Kummerfeld, Zhang,
  Ramanathan, Sadasivam, Zhang, and Radev}]{finegan2018improving}
Catherine Finegan-Dollak, Jonathan~K Kummerfeld, Li~Zhang, Karthik Ramanathan,
  Sesh Sadasivam, Rui Zhang, and Dragomir Radev. 2018.
\newblock Improving text-to-sql evaluation methodology.
\newblock \emph{arXiv preprint arXiv:1806.09029}.

\bibitem[{Guo et~al.(2020)Guo, Liu, Lou, Li, Liu, Xie, and
  Liu}]{guo2020benchmarking}
Jiaqi Guo, Qian Liu, Jian-Guang Lou, Zhenwen Li, Xueqing Liu, Tao Xie, and Ting
  Liu. 2020.
\newblock Benchmarking meaning representations in neural semantic parsing.
\newblock In \emph{Proceedings of the 2020 Conference on Empirical Methods in
  Natural Language Processing (EMNLP)}, pages 1520--1540.

\bibitem[{Gupta et~al.(2018)Gupta, Shah, Mohit, Kumar, and
  Lewis}]{gupta2018semantic}
Sonal Gupta, Rushin Shah, Mrinal Mohit, Anuj Kumar, and Mike Lewis. 2018.
\newblock Semantic parsing for task oriented dialog using hierarchical
  representations.
\newblock \emph{arXiv preprint arXiv:1810.07942}.

\bibitem[{Haas and Riezler(2016)}]{haas2016corpus}
Carolin Haas and Stefan Riezler. 2016.
\newblock A corpus and semantic parser for multilingual natural language
  querying of openstreetmap.
\newblock In \emph{Proceedings of the 2016 Conference of the North American
  Chapter of the Association for Computational Linguistics: Human Language
  Technologies}, pages 740--750.

\bibitem[{Han et~al.(2022)Han, Schoelkopf, Zhao, Qi, Riddell, Benson, Sun,
  Zubova, Qiao, Burtell et~al.}]{han2022folio}
Simeng Han, Hailey Schoelkopf, Yilun Zhao, Zhenting Qi, Martin Riddell, Luke
  Benson, Lucy Sun, Ekaterina Zubova, Yujie Qiao, Matthew Burtell, et~al. 2022.
\newblock Folio: Natural language reasoning with first-order logic.
\newblock \emph{arXiv preprint arXiv:2209.00840}.

\bibitem[{Hu et~al.(2020)Hu, Ruder, Siddhant, Neubig, Firat, and
  Johnson}]{hu2020xtreme}
Junjie Hu, Sebastian Ruder, Aditya Siddhant, Graham Neubig, Orhan Firat, and
  Melvin Johnson. 2020.
\newblock Xtreme: A massively multilingual multi-task benchmark for evaluating
  cross-lingual generalisation.
\newblock In \emph{International Conference on Machine Learning}, pages
  4411--4421. PMLR.

\bibitem[{Iyer et~al.(2017)Iyer, Konstas, Cheung, Krishnamurthy, and
  Zettlemoyer}]{iyer2017learning}
Srinivasan Iyer, Ioannis Konstas, Alvin Cheung, Jayant Krishnamurthy, and Luke
  Zettlemoyer. 2017.
\newblock Learning a neural semantic parser from user feedback.
\newblock \emph{arXiv preprint arXiv:1704.08760}.

\bibitem[{Jones et~al.(2012)Jones, Johnson, and Goldwater}]{jones2012semantic}
Bevan Jones, Mark Johnson, and Sharon Goldwater. 2012.
\newblock Semantic parsing with bayesian tree transducers.
\newblock In \emph{Proceedings of the 50th Annual Meeting of the Association
  for Computational Linguistics (Volume 1: Long Papers)}, pages 488--496.

\bibitem[{Keysers et~al.(2020)Keysers, Sch{\"{a}}rli, Scales, Buisman, Furrer,
  Kashubin, Momchev, Sinopalnikov, Stafiniak, Tihon, Tsarkov, Wang, van Zee,
  and Bousquet}]{DBLP:conf/iclr/KeysersSSBFKMSS20}
Daniel Keysers, Nathanael Sch{\"{a}}rli, Nathan Scales, Hylke Buisman, Daniel
  Furrer, Sergii Kashubin, Nikola Momchev, Danila Sinopalnikov, Lukasz
  Stafiniak, Tibor Tihon, Dmitry Tsarkov, Xiao Wang, Marc van Zee, and Olivier
  Bousquet. 2020.
\newblock \href {https://openreview.net/forum?id=SygcCnNKwr} {Measuring
  compositional generalization: {A} comprehensive method on realistic data}.
\newblock In \emph{8th International Conference on Learning Representations,
  {ICLR} 2020, Addis Ababa, Ethiopia, April 26-30, 2020}. OpenReview.net.

\bibitem[{Keysers et~al.(2019)Keysers, Sch{\"a}rli, Scales, Buisman, Furrer,
  Kashubin, Momchev, Sinopalnikov, Stafiniak, Tihon
  et~al.}]{keysers2019measuring}
Daniel Keysers, Nathanael Sch{\"a}rli, Nathan Scales, Hylke Buisman, Daniel
  Furrer, Sergii Kashubin, Nikola Momchev, Danila Sinopalnikov, Lukasz
  Stafiniak, Tibor Tihon, et~al. 2019.
\newblock Measuring compositional generalization: A comprehensive method on
  realistic data.
\newblock \emph{arXiv preprint arXiv:1912.09713}.

\bibitem[{Lample and Conneau(2019)}]{lample2019cross}
Guillaume Lample and Alexis Conneau. 2019.
\newblock Cross-lingual language model pretraining.
\newblock \emph{Advances in Neural Information Processing Systems (NeurIPS)}.

\bibitem[{Lauren{\c{c}}on et~al.(2022)Lauren{\c{c}}on, Saulnier, Wang, Akiki,
  del Moral, Le~Scao, Von~Werra, Mou, Ponferrada, Nguyen
  et~al.}]{laurenccon2022bigscience}
Hugo Lauren{\c{c}}on, Lucile Saulnier, Thomas Wang, Christopher Akiki,
  Albert~Villanova del Moral, Teven Le~Scao, Leandro Von~Werra, Chenghao Mou,
  Eduardo~Gonz{\'a}lez Ponferrada, Huu Nguyen, et~al. 2022.
\newblock The bigscience roots corpus: A 1.6 tb composite multilingual dataset.
\newblock In \emph{Thirty-sixth Conference on Neural Information Processing
  Systems Datasets and Benchmarks Track}.

\bibitem[{Lawrence and Riezler(2016)}]{lawrence2016nlmaps}
Carolin Lawrence and Stefan Riezler. 2016.
\newblock Nlmaps: A natural language interface to query openstreetmap.
\newblock In \emph{Proceedings of COLING 2016, the 26th International
  Conference on Computational Linguistics: System Demonstrations}, pages 6--10.

\bibitem[{Lewis et~al.(2019)Lewis, O{\u{g}}uz, Rinott, Riedel, and
  Schwenk}]{lewis2019mlqa}
Patrick Lewis, Barlas O{\u{g}}uz, Ruty Rinott, Sebastian Riedel, and Holger
  Schwenk. 2019.
\newblock Mlqa: Evaluating cross-lingual extractive question answering.
\newblock \emph{arXiv preprint arXiv:1910.07475}.

\bibitem[{Li et~al.(2020)Li, Arora, Chen, Gupta, Gupta, and
  Mehdad}]{li2020mtop}
Haoran Li, Abhinav Arora, Shuohui Chen, Anchit Gupta, Sonal Gupta, and Yashar
  Mehdad. 2020.
\newblock Mtop: A comprehensive multilingual task-oriented semantic parsing
  benchmark.
\newblock \emph{arXiv preprint arXiv:2008.09335}.

\bibitem[{Li et~al.(2023)Li, Hui, Cheng, Qin, Ma, Huo, Huang, Du, Si, and
  Li}]{li2023graphix}
Jinyang Li, Binyuan Hui, Reynold Cheng, Bowen Qin, Chenhao Ma, Nan Huo, Fei
  Huang, Wenyu Du, Luo Si, and Yongbin Li. 2023.
\newblock Graphix-t5: Mixing pre-trained transformers with graph-aware layers
  for text-to-sql parsing.
\newblock \emph{arXiv preprint arXiv:2301.07507}.

\bibitem[{Li et~al.(2022)Li, Guo, Liu, Lou, and
  Xie}]{li-etal-2022-exploring-secrets}
Zhenwen Li, Jiaqi Guo, Qian Liu, Jian-Guang Lou, and Tao Xie. 2022.
\newblock \href {https://aclanthology.org/2022.emnlp-main.237} {Exploring the
  secrets behind the learning difficulty of meaning representations for
  semantic parsing}.
\newblock In \emph{Proceedings of the 2022 Conference on Empirical Methods in
  Natural Language Processing}, pages 3616--3625, Abu Dhabi, United Arab
  Emirates. Association for Computational Linguistics.

\bibitem[{Liang et~al.(2020)Liang, Duan, Gong, Wu, Guo, Qi, Gong, Shou, Jiang,
  Cao et~al.}]{liang2020xglue}
Yaobo Liang, Nan Duan, Yeyun Gong, Ning Wu, Fenfei Guo, Weizhen Qi, Ming Gong,
  Linjun Shou, Daxin Jiang, Guihong Cao, et~al. 2020.
\newblock Xglue: A new benchmark dataset for cross-lingual pre-training,
  understanding and generation.
\newblock \emph{arXiv preprint arXiv:2004.01401}.

\bibitem[{Lin et~al.(2022)Lin, Mihaylov, Artetxe, Wang, Chen, Simig, Ott,
  Goyal, Bhosale, Du et~al.}]{lin2022few}
Xi~Victoria Lin, Todor Mihaylov, Mikel Artetxe, Tianlu Wang, Shuohui Chen,
  Daniel Simig, Myle Ott, Naman Goyal, Shruti Bhosale, Jingfei Du, et~al. 2022.
\newblock Few-shot learning with multilingual generative language models.
\newblock In \emph{Proceedings of the 2022 Conference on Empirical Methods in
  Natural Language Processing}, pages 9019--9052.

\bibitem[{Liu et~al.(2020)Liu, Gu, Goyal, Li, Edunov, Ghazvininejad, Lewis, and
  Zettlemoyer}]{liu2020multilingual}
Yinhan Liu, Jiatao Gu, Naman Goyal, Xian Li, Sergey Edunov, Marjan
  Ghazvininejad, Mike Lewis, and Luke Zettlemoyer. 2020.
\newblock Multilingual denoising pre-training for neural machine translation.
\newblock \emph{Transactions of the Association for Computational Linguistics},
  8:726--742.

\bibitem[{Liu et~al.(2019)Liu, Ott, Goyal, Du, Joshi, Chen, Levy, Lewis,
  Zettlemoyer, and Stoyanov}]{liu2019roberta}
Yinhan Liu, Myle Ott, Naman Goyal, Jingfei Du, Mandar Joshi, Danqi Chen, Omer
  Levy, Mike Lewis, Luke Zettlemoyer, and Veselin Stoyanov. 2019.
\newblock Roberta: A robustly optimized bert pretraining approach.
\newblock \emph{arXiv preprint arXiv:1907.11692}.

\bibitem[{Lu and Ng(2011)}]{lu2011probabilistic}
Wei Lu and Hwee~Tou Ng. 2011.
\newblock A probabilistic forest-to-string model for language generation from
  typed lambda calculus expressions.
\newblock In \emph{Proceedings of the 2011 Conference on Empirical Methods in
  Natural Language Processing}, pages 1611--1622.

\bibitem[{Min et~al.(2019)Min, Shi, and Zhang}]{min2019pilot}
Qingkai Min, Yuefeng Shi, and Yue Zhang. 2019.
\newblock A pilot study for chinese sql semantic parsing.
\newblock \emph{arXiv preprint arXiv:1909.13293}.

\bibitem[{Moradshahi et~al.(2020)Moradshahi, Campagna, Semnani, Xu, and
  Lam}]{moradshahi2020localizing}
Mehrad Moradshahi, Giovanni Campagna, Sina~J Semnani, Silei Xu, and Monica~S
  Lam. 2020.
\newblock Localizing open-ontology qa semantic parsers in a day using machine
  translation.
\newblock \emph{arXiv preprint arXiv:2010.05106}.

\bibitem[{Nguyen et~al.(2020)Nguyen, Dao, and Nguyen}]{nguyen2020pilot}
Anh~Tuan Nguyen, Mai~Hoang Dao, and Dat~Quoc Nguyen. 2020.
\newblock A pilot study of text-to-sql semantic parsing for vietnamese.
\newblock \emph{arXiv preprint arXiv:2010.01891}.

\bibitem[{Oard et~al.(2019)Oard, Carpuat, Galu{\v{s}}{\v{c}}{\'a}kov{\'a},
  Barrow, Nair, Niu, Shing, Xu, Zotkina, McKeown et~al.}]{oard2019surprise}
Douglas~W Oard, Marine Carpuat, Petra Galu{\v{s}}{\v{c}}{\'a}kov{\'a}, Joseph
  Barrow, Suraj Nair, Xing Niu, Han-Chin Shing, Weijia Xu, Elena Zotkina,
  Kathleen McKeown, et~al. 2019.
\newblock Surprise languages: rapid-response cross-language ir.
\newblock In \emph{ACM NTCIR-14 Conference}, volume~10.

\bibitem[{Pfeiffer et~al.(2022)Pfeiffer, Goyal, Lin, Li, Cross, Riedel, and
  Artetxe}]{pfeiffer-etal-2022-lifting}
Jonas Pfeiffer, Naman Goyal, Xi~Lin, Xian Li, James Cross, Sebastian Riedel,
  and Mikel Artetxe. 2022.
\newblock \href {https://doi.org/10.18653/v1/2022.naacl-main.255} {Lifting the
  curse of multilinguality by pre-training modular transformers}.
\newblock In \emph{Proceedings of the 2022 Conference of the North American
  Chapter of the Association for Computational Linguistics: Human Language
  Technologies}, pages 3479--3495, Seattle, United States. Association for
  Computational Linguistics.

\bibitem[{Prakash et~al.(2020)Prakash, Shashidhar, Zhao, Rongali, Khan, and
  Kayser}]{prakash2020compressing}
Prafull Prakash, Saurabh~Kumar Shashidhar, Wenlong Zhao, Subendhu Rongali,
  Haidar Khan, and Michael Kayser. 2020.
\newblock Compressing transformer-based semantic parsing models using
  compositional code embeddings.
\newblock \emph{arXiv preprint arXiv:2010.05002}.

\bibitem[{Price(1990)}]{price1990evaluation}
Patti Price. 1990.
\newblock Evaluation of spoken language systems: The atis domain.
\newblock In \emph{Speech and Natural Language: Proceedings of a Workshop Held
  at Hidden Valley, Pennsylvania, June 24-27, 1990}.

\bibitem[{Rongali et~al.(2020)Rongali, Soldaini, Monti, and
  Hamza}]{rongali2020don}
Subendhu Rongali, Luca Soldaini, Emilio Monti, and Wael Hamza. 2020.
\newblock Don’t parse, generate! a sequence to sequence architecture for
  task-oriented semantic parsing.
\newblock In \emph{Proceedings of The Web Conference 2020}, pages 2962--2968.

\bibitem[{Scao et~al.(2022)Scao, Fan, Akiki, Pavlick, Ili{\'c}, Hesslow,
  Castagn{\'e}, Luccioni, Yvon, Gall{\'e} et~al.}]{scao2022bloom}
Teven~Le Scao, Angela Fan, Christopher Akiki, Ellie Pavlick, Suzana Ili{\'c},
  Daniel Hesslow, Roman Castagn{\'e}, Alexandra~Sasha Luccioni, Fran{\c{c}}ois
  Yvon, Matthias Gall{\'e}, et~al. 2022.
\newblock Bloom: A 176b-parameter open-access multilingual language model.
\newblock \emph{arXiv preprint arXiv:2211.05100}.

\bibitem[{Sherborne and Lapata(2021)}]{sherborne2021zero}
Tom Sherborne and Mirella Lapata. 2021.
\newblock Zero-shot cross-lingual semantic parsing.
\newblock \emph{arXiv preprint arXiv:2104.07554}.

\bibitem[{Sherborne and Lapata(2022)}]{sherborne-lapata-2022-zero}
Tom Sherborne and Mirella Lapata. 2022.
\newblock \href {https://doi.org/10.18653/v1/2022.acl-long.285} {Zero-shot
  cross-lingual semantic parsing}.
\newblock In \emph{Proceedings of the 60th Annual Meeting of the Association
  for Computational Linguistics (Volume 1: Long Papers)}, pages 4134--4153,
  Dublin, Ireland. Association for Computational Linguistics.

\bibitem[{Sherborne and Lapata(2023)}]{sherborne2023meta}
Tom Sherborne and Mirella Lapata. 2023.
\newblock Meta-learning a cross-lingual manifold for semantic parsing.
\newblock \emph{Transactions of the Association for Computational Linguistics},
  11:49--67.

\bibitem[{Sherborne et~al.(2020)Sherborne, Xu, and
  Lapata}]{sherborne2020bootstrapping}
Tom Sherborne, Yumo Xu, and Mirella Lapata. 2020.
\newblock Bootstrapping a crosslingual semantic parser.
\newblock \emph{arXiv preprint arXiv:2004.02585}.

\bibitem[{Shi et~al.(2022{\natexlab{a}})Shi, Song, Jin, Mi, Bai, Lin, and
  Yu}]{shi-etal-2022-cross}
Peng Shi, Linfeng Song, Lifeng Jin, Haitao Mi, He~Bai, Jimmy Lin, and Dong Yu.
  2022{\natexlab{a}}.
\newblock \href {https://aclanthology.org/2022.findings-emnlp.388}
  {Cross-lingual text-to-{SQL} semantic parsing with representation mixup}.
\newblock In \emph{Findings of the Association for Computational Linguistics:
  EMNLP 2022}, pages 5296--5306, Abu Dhabi, United Arab Emirates. Association
  for Computational Linguistics.

\bibitem[{Shi et~al.(2021)Shi, Zhang, Bai, and Lin}]{shi2021cross}
Peng Shi, Rui Zhang, He~Bai, and Jimmy Lin. 2021.
\newblock Cross-lingual training of dense retrievers for document retrieval.
\newblock In \emph{Proceedings of the 1st Workshop on Multilingual
  Representation Learning}, pages 251--253.

\bibitem[{Shi et~al.(2022{\natexlab{b}})Shi, Zhang, Bai, and
  Lin}]{shi2022xricl}
Peng Shi, Rui Zhang, He~Bai, and Jimmy Lin. 2022{\natexlab{b}}.
\newblock Xricl: Cross-lingual retrieval-augmented in-context learning for
  cross-lingual text-to-sql semantic parsing.
\newblock \emph{arXiv preprint arXiv:2210.13693}.

\bibitem[{Shu et~al.(2021)Shu, Zhang, Dong, Shi, Yu, and
  Zhang}]{shu-etal-2021-logic}
Chang Shu, Yusen Zhang, Xiangyu Dong, Peng Shi, Tao Yu, and Rui Zhang. 2021.
\newblock \href {https://doi.org/10.18653/v1/2021.findings-acl.388}
  {Logic-consistency text generation from semantic parses}.
\newblock In \emph{Findings of the Association for Computational Linguistics:
  ACL-IJCNLP 2021}, pages 4414--4426, Online. Association for Computational
  Linguistics.

\bibitem[{Srivastava et~al.(2022)Srivastava, Rastogi, Rao, Shoeb, Abid, Fisch,
  Brown, Santoro, Gupta, Garriga-Alonso et~al.}]{srivastava2022beyond}
Aarohi Srivastava, Abhinav Rastogi, Abhishek Rao, Abu Awal~Md Shoeb, Abubakar
  Abid, Adam Fisch, Adam~R Brown, Adam Santoro, Aditya Gupta, Adri{\`a}
  Garriga-Alonso, et~al. 2022.
\newblock Beyond the imitation game: Quantifying and extrapolating the
  capabilities of language models.
\newblock \emph{arXiv preprint arXiv:2206.04615}.

\bibitem[{Sun and Duh(2020)}]{sun2020clirmatrix}
Shuo Sun and Kevin Duh. 2020.
\newblock Clirmatrix: A massively large collection of bilingual and
  multilingual datasets for cross-lingual information retrieval.
\newblock In \emph{Proceedings of the 2020 Conference on Empirical Methods in
  Natural Language Processing (EMNLP)}, pages 4160--4170.

\bibitem[{Susanto and Lu(2017{\natexlab{a}})}]{susanto2017neural}
Raymond~Hendy Susanto and Wei Lu. 2017{\natexlab{a}}.
\newblock Neural architectures for multilingual semantic parsing.
\newblock In \emph{Proceedings of the 55th Annual Meeting of the Association
  for Computational Linguistics (Volume 2: Short Papers)}, pages 38--44.

\bibitem[{Susanto and Lu(2017{\natexlab{b}})}]{susanto2017semantic}
Raymond~Hendy Susanto and Wei Lu. 2017{\natexlab{b}}.
\newblock Semantic parsing with neural hybrid trees.
\newblock In \emph{Proceedings of the AAAI Conference on Artificial
  Intelligence}.

\bibitem[{Upadhyay et~al.(2018)Upadhyay, Faruqui, T{\"u}r, Dilek, and
  Heck}]{upadhyay2018almost}
Shyam Upadhyay, Manaal Faruqui, Gokhan T{\"u}r, Hakkani-T{\"u}r Dilek, and
  Larry Heck. 2018.
\newblock (almost) zero-shot cross-lingual spoken language understanding.
\newblock In \emph{2018 IEEE International Conference on Acoustics, Speech and
  Signal Processing (ICASSP)}, pages 6034--6038. IEEE.

\bibitem[{Vaswani et~al.(2017)Vaswani, Shazeer, Parmar, Uszkoreit, Jones,
  Gomez, Kaiser, and Polosukhin}]{vaswani2017attention}
Ashish Vaswani, Noam Shazeer, Niki Parmar, Jakob Uszkoreit, Llion Jones,
  Aidan~N Gomez, Lukasz Kaiser, and Illia Polosukhin. 2017.
\newblock Attention is all you need.
\newblock \emph{arXiv preprint arXiv:1706.03762}.

\bibitem[{Wang et~al.(2015)Wang, Berant, and Liang}]{wang2015building}
Yushi Wang, Jonathan Berant, and Percy Liang. 2015.
\newblock Building a semantic parser overnight.
\newblock In \emph{Proceedings of the 53rd Annual Meeting of the Association
  for Computational Linguistics and the 7th International Joint Conference on
  Natural Language Processing (Volume 1: Long Papers)}, pages 1332--1342.

\bibitem[{Wang et~al.(2022)Wang, Cuenca, Zhou, Xu, and
  Neubig}]{wang2022mconala}
Zhiruo Wang, Grace Cuenca, Shuyan Zhou, Frank~F Xu, and Graham Neubig. 2022.
\newblock Mconala: A benchmark for code generation from multiple natural
  languages.
\newblock \emph{arXiv preprint arXiv:2203.08388}.

\bibitem[{Wu et~al.(2016)Wu, Schuster, Chen, Le, Norouzi, Macherey, Krikun,
  Cao, Gao, Macherey et~al.}]{wu2016google}
Yonghui Wu, Mike Schuster, Zhifeng Chen, Quoc~V Le, Mohammad Norouzi, Wolfgang
  Macherey, Maxim Krikun, Yuan Cao, Qin Gao, Klaus Macherey, et~al. 2016.
\newblock Google's neural machine translation system: Bridging the gap between
  human and machine translation.
\newblock \emph{arXiv preprint arXiv:1609.08144}.

\bibitem[{Xu et~al.(2020{\natexlab{a}})Xu, Campagna, Li, and
  Lam}]{xu2020schema2qa}
Silei Xu, Giovanni Campagna, Jian Li, and Monica~S Lam. 2020{\natexlab{a}}.
\newblock Schema2qa: High-quality and low-cost q\&a agents for the structured
  web.
\newblock In \emph{Proceedings of the 29th ACM International Conference on
  Information \& Knowledge Management}, pages 1685--1694.

\bibitem[{Xu et~al.(2020{\natexlab{b}})Xu, Haider, and Mansour}]{xu2020end}
Weijia Xu, Batool Haider, and Saab Mansour. 2020{\natexlab{b}}.
\newblock End-to-end slot alignment and recognition for cross-lingual nlu.
\newblock \emph{arXiv preprint arXiv:2004.14353}.

\bibitem[{Xue et~al.(2020)Xue, Constant, Roberts, Kale, Al-Rfou, Siddhant,
  Barua, and Raffel}]{xue2020mt5}
Linting Xue, Noah Constant, Adam Roberts, Mihir Kale, Rami Al-Rfou, Aditya
  Siddhant, Aditya Barua, and Colin Raffel. 2020.
\newblock \href {http://arxiv.org/abs/2010.11934} {{mT5}: A massively
  multilingual pre-trained text-to-text transformer}.

\bibitem[{Yin et~al.(2018)Yin, Deng, Chen, Vasilescu, and
  Neubig}]{yin2018learning}
Pengcheng Yin, Bowen Deng, Edgar Chen, Bogdan Vasilescu, and Graham Neubig.
  2018.
\newblock Learning to mine aligned code and natural language pairs from stack
  overflow.
\newblock In \emph{2018 IEEE/ACM 15th international conference on mining
  software repositories (MSR)}, pages 476--486. IEEE.

\bibitem[{Yu et~al.(2018)Yu, Zhang, Yang, Yasunaga, Wang, Li, Ma, Li, Yao,
  Roman et~al.}]{yu2018spider}
Tao Yu, Rui Zhang, Kai Yang, Michihiro Yasunaga, Dongxu Wang, Zifan Li, James
  Ma, Irene Li, Qingning Yao, Shanelle Roman, et~al. 2018.
\newblock Spider: A large-scale human-labeled dataset for complex and
  cross-domain semantic parsing and text-to-sql task.
\newblock \emph{arXiv preprint arXiv:1809.08887}.

\bibitem[{Zbib et~al.(2019)Zbib, Zhao, Karakos, Hartmann, DeYoung, Huang,
  Jiang, Rivkin, Zhang, Schwartz et~al.}]{zbib2019neural}
Rabih Zbib, Lingjun Zhao, Damianos Karakos, William Hartmann, Jay DeYoung,
  Zhongqiang Huang, Zhuolin Jiang, Noah Rivkin, Le~Zhang, Richard Schwartz,
  et~al. 2019.
\newblock Neural-network lexical translation for cross-lingual ir from text and
  speech.
\newblock In \emph{Proceedings of the 42nd International ACM SIGIR Conference
  on Research and Development in Information Retrieval}, pages 645--654.

\bibitem[{Zelle and Mooney(1996)}]{zelle1996learning}
John~M Zelle and Raymond~J Mooney. 1996.
\newblock Learning to parse database queries using inductive logic programming.
\newblock In \emph{Proceedings of the national conference on artificial
  intelligence}, pages 1050--1055.

\bibitem[{Zettlemoyer and Collins(2012)}]{zettlemoyer2012learning}
Luke~S Zettlemoyer and Michael Collins. 2012.
\newblock Learning to map sentences to logical form: Structured classification
  with probabilistic categorial grammars.
\newblock \emph{arXiv preprint arXiv:1207.1420}.

\bibitem[{Zhang et~al.(2019)Zhang, Westerfield, Shim, Bingham, Fabbri, Verma,
  Hu, and Radev}]{zhang2019improving}
Rui Zhang, Caitlin Westerfield, Sungrok Shim, Garrett Bingham, Alexander
  Fabbri, Neha Verma, William Hu, and Dragomir Radev. 2019.
\newblock Improving low-resource cross-lingual document retrieval by reranking
  with deep bilingual representations.
\newblock \emph{arXiv preprint arXiv:1906.03492}.

\bibitem[{Zhu et~al.(2019)Zhu, Wang, Wang, Zhou, Zhang, Wang, and
  Zong}]{zhu2019ncls}
Junnan Zhu, Qian Wang, Yining Wang, Yu~Zhou, Jiajun Zhang, Shaonan Wang, and
  Chengqing Zong. 2019.
\newblock Ncls: Neural cross-lingual summarization.
\newblock \emph{arXiv preprint arXiv:1909.00156}.

\bibitem[{Zou and Lu(2018)}]{zou2018learning}
Yanyan Zou and Wei Lu. 2018.
\newblock Learning cross-lingual distributed logical representations for
  semantic parsing.
\newblock \emph{arXiv preprint arXiv:1806.05461}.

\end{thebibliography}
\bibliographystyle{acl_natbib}

\appendix
\label{sec:appendix}

\section{Data Construction Details}
In this section, we introduce the details of data collection, natural languages, meaning representations, and dataset statistics.

\subsection*{A.1 \quad Data Collection}
\label{sec:app:construction}
\paragraph{Multilingual ATIS}
ATIS~\cite{price1990evaluation,dahl1994expanding} contains user questions for a flight-booking task.
The original user questions are in English.
We add the translations in Spanish, German, French, Portuguese, Japanese, Chinese from~\citet{xu2020end}.
Furthermore, \citet{upadhyay2018almost} provide translations in Hindi and Turkish but only for a subset of utterances.
\citet{susanto2017neural} provide translations in Indonesian and Chinese, and \citet{sherborne2020bootstrapping} provide translations in Chinese and German, but neither is available through LDC.
Therefore, we don't include these.
For meaning representations, we focus on the task of NLI to databases and thus collect SQL from~\citet{iyer2017learning,finegan2018improving}, while there are other formats available such as logical forms~\cite{zettlemoyer2012learning} and BIO tags for slot and intent~\cite{upadhyay2018almost}.
To unify SQL formats across datasets, we rewrite the SQL queries following the format of Spider~\cite{yu2018spider}.
We follow the question splits from~\citet{finegan2018improving}.
Through manual inspection, we discard 52 examples which do not have aligned translations from~\citet{xu2020end}.
This gives 5228 examples with 4303 training, 481 dev, and 444 test.

\paragraph{Multilingual GeoQuery}
GeoQuery~\cite{zelle1996learning} contains user questions about US geography. 
The original user questions are in English.
One of the earliest work on cross-lingual semantic parsing is on the Chinese version of GeoQuery created by~\citet{lu2011probabilistic}.
Later, \citet{jones2012semantic} create German, Greek, and Thai translations, and~\citet{susanto2017semantic} create Indonesian, Swedish, and Farsi translations.
We include all these 8 languages.
Furthermore, GeoQuery has several meaning representations available.
To include multiple meaning representations, we collect Prolog and Lambda Calculus from~\citet{guo2020benchmarking}, FunQL from~\citet{susanto2017semantic}, and SQL from~\citet{finegan2018improving}.
To unify SQL formats across datasets, we rewrite the SQL queries following the format of Spider~\cite{yu2018spider}.
We follow the question splits from~\citet{finegan2018improving}.
Through manual inspection, we discard 3 examples that do not have corresponding FunQL representations.
This gives 874 examples with 548 training, 49 dev, and 277 test.

\paragraph{Multilingual Spider}
Spider~\cite{yu2018spider} is a human-annotated complex and cross-domain text-to-SQL datasets.
The original Spider uses English utterances and database schemas. 
To include utterances in other languages, we include the Chinese version~\cite{min2019pilot} and the syllable-level Vietnamese version~\cite{nguyen2020pilot}.
In this way, each SQL query is paired with a database schema in English and an utterance in three languages.
Because the test set is not public, we include only the training and dev set.
We also exclude GeoQuery examples from its training set because we use the full version of GeoQuery separately.
This creates 8095 training examples and 1034 dev examples following the original splits~\cite{yu2018spider}.

\paragraph{Multilingual NLmaps}
NLMaps~\cite{lawrence2016nlmaps} is a Natural Language Interface to query the OpenStreetMap database about geographical facts.
The original questions are in English, and later~\citet{haas2016corpus} provide translations in German.
The meaning representation is Functional Query Language designed for OpenStreetMap, which is similar to FunQL of GeoQuery.
We follow the original split with 1500 training and 880 test examples.

\paragraph{Multilingual Overnight}
Overnight~\cite{wang2015building} is a multi-domain semantic parsing dataset with lambda DCS logical forms executable in SEMPRE~\cite{berant2013semantic}.
The questions cover 8 domains in Calendar, Blocks, Housing, Restaurants, Recipes, Publications, Social, Basketball.
The original dataset is in English, and~\citet{sherborne2020bootstrapping} provide translation in German and Chinese.
They use machine translation for the training set and human translation on the dev and test sets.
We include the Baidu Translation for Chinese and Google Translate for German.
We merge all the domains together as a single dataset and follow the original split with 8754 training, 2188 dev, and 2740 test examples.

\paragraph{MCWQ}
MCWQ~\cite{cui2021multilingual} is a multilingual knowledge-based question answering dataset grounded in Wikidata.
This is created by adapting the CFQ (Compositional Freebase Questions) dataset~\cite{keysers2019measuring} by translating the queries into SQARQL for Wikidata.
The questions are in four languages including Hebrew, Kannada, Chinese, and English.
The split follows maximum compound divergence (MCD) so that the test set contains novel compounds to test compositionality generalization ability.
We follow the MCD3 splits with 4006 training, 733 dev, and 648 test examples.

\paragraph{Multilingual Schema2QA}
Schema2QA~\cite{xu2020schema2qa} is an open-ontology question answering dataset over scraped Schema.org web data with meaning representations in ThingTalk Query Language.
\citet{moradshahi2020localizing} extend the original dataset with utterances in English, Arabic, German, Spanish, Farsi, Finnish, Italian, Japanese, Polish, Turkish, Chinese.
The questions cover 2 domains in hotels and restaurants.
The training examples are automatically generated based on template-based synthesis, crowdsourced paraphrasing, and machine translation.
The test examples are crowdsourced and manually annotated by an expert with human translations.
We include training examples with all 11 languages available and pair the translations with the query in corresponding language.
To make the dataset size comparable to others, we include 5\% of the training set.
This gives 8932 training examples and 971 test examples.
We also include a no-value version of the query, because the entities in the translated utterances are localized in the new languages and thus do not align well with the values in English queries.

\paragraph{MTOP}
MTOP~\cite{li2020mtop} is a multilingual task-oriented semantic parsing dataset with meaning representations based on hierarchical intent and slot annotations~\cite{gupta2018semantic}.
It covers 11 domains in Alarm, Calling, Event, Messaging, Music, News, People, Recipes, Reminder, Timer, Weather.
It includes 6 languages in English, German, French, Spanish, Hindi, Thai.
We include examples with all 6 languages available and pair the translations with the compositional decoupled representation in corresponding language.
This gives 5446 training, 863 dev, 1245 test examples.

\paragraph{MCoNaLa} MCoNaLa~\citep{wang2022mconala} is a code generation benchmark which requires to generate Python code. It collects English examples from the English Code/Natural Language Challenge (CoNaLa~\citep{yin2018learning}) dataset and further annotates a total of 896 NL-code pairs in three languages, including Spanish, Japanese, and Russian. The training and dev set contains 1903 and 476 English examples, separately.

\begin{table*}[t!]
\centering
\resizebox{\textwidth}{!}{%
\begin{tabular}{p{0.1\linewidth}  p{0.2\linewidth} p{0.8\linewidth}}
\hline
Dataset   & Utterance & Meaning Representation (MR) \\ \hline
ATIS      &  Liste todos os voos que chegam ao General Mitchell International de várias cidades & SELECT DISTINCT T3.FLIGHT\_ID FROM CITY AS T1 JOIN AIRPORT\_SERVICE AS T2 ON T1.CITY\_CODE = T2.CITY\_CODE JOIN FLIGHT AS T3 ON T3.FROM\_AIRPORT = T2.AIRPORT\_CODE JOIN AIRPORT AS T4 ON T3.TO\_AIRPORT = T4.AIRPORT\_CODE WHERE T4.AIRPORT\_CODE = ``MKE" \\
GeoQuery  &  \foreignlanguage{farsi}{بزرگترین شهر لوئیزیانا کدام است ؟} & answer(A,largest(A,(city(A),loc(A,B),const(B,stateid(louisiana)))))  \\   
Spider    &  \begin{CJK}{UTF8}{gbsn}有多少摇滚歌手\end{CJK} &  SELECT count(*) FROM singer WHERE genre = `Rock'                           \\
NLmaps    &     Wie viele japanische Restaurants gibt es in Paris?      & query(area(keyval(`name',`Paris'), keyval(`is\_in:country',`France')), nwr(keyval(`cuisine',`japanese')),qtype(count))    \\
Overnight &     what players made less than three assists over a season      & ( call SW.listValue ( call SW.getProperty ( ( lambda s ( call SW.filter ( var s ) ( call SW.ensureNumericProperty ( string num\_assists ) ) ( string < ) ( call SW.ensureNumericEntity ( number 3 assist ) ) ) ) ( call SW.domain ( string player ) ) ) ( string player ) ) )                            \\
MCWQ & \quad \<ht.htN> M0 \<h'M hyld /sl> \flushright{M1 \<`M>}   & ASK WHERE { ?x0 wdt:P749 M0 . ?x0 wdt:P26 M1 . FILTER ( ?x0 != M1 ) }\\
Schema2QA & \selectlanguage{italian}{mostrami i ristoranti con pi\`u recensioni}  & now => ( sort param:aggregateRating.reviewCount:Number desc of ( @org.schema.Restaurant.Restaurant ) ) [ 1 ] => notify\\
MTOP & {\dn pMpEkn rn Eks \3FEwkAr kA iv\?{\qva}V h\4 {\rs ?\re}} &  [IN:GET\_CATEGORY\_EVENT [SL:TITLE\_EVENT {\dn pMpEkn rn} ] ] \\

MCoNaLa & \begin{CJK}{UTF8}{min}タプルdataを空白区切りで表示する\end{CJK} & for i in data:
     print(' '.join(str(j) for j in i)) \\
\hline
\end{tabular}
}
\caption{Examples of each dataset in \ours including diverse languages and meaning representations.
ATIS: Portuguese-SQL, Geoquery: Farsi-Prolog, Spider: Vietnamese-SQL, NLmaps: German-FunQL, Overnight: English-Lambda Calculus, MCWQ: Hebrew-SPARQL, Schema2QA: Arabic-ThingTalk Query Language, MTOP: Hindi-Hierarchical Intent and Slot, MCoNaLa: Japanese-Python.}
\label{tab:example}
\end{table*}

\subsection*{A.2 \quad Language Details}
\label{sec:app:language}
We assemble 9 datasets in various domains for 5 semantic parsing tasks. 
It covers 8 meaning representations: SQL, Lambda Calculus, Functional Query Language, Prolog, SPARQL, ThingTalk Query Language, Python, Hierarchical Intent and Slot.
The questions covers 22 languages in 15 language families: Arabic(Afro-Asiatic), Chinese(Sino-Tibetan), English(IE: Germanic), Farsi(IE: Iranian), Finnish(Uralic), French(IE: Romance), German(IE: Germanic), Greek(IE: Greek), Hebrew(Afro-Asiatic), Hindi(IE: Indo-Aryan), Indonesian(Austronesian), Italian(IE: Romance), Japanese(Japonic), Kannada(Dravidian), Polish(IE: Slavic), Portuguese(IE: Romance), Russian(IE: Slavic), Spanish(IE: Romance), Swedish(IE: Germanic), Thai(Kra-Dai), Turkish(Turkic), Vietnamese(Austro-Asiatic). Each dataset has English for cross-lingual transfer over other languages.

\subsection*{A.3 \quad Meaning Representation Details}
\label{sec:app:mr}
Prolog uses first-order logic augmented with higher-order predicates for quantification and aggregation. 
Lambda Calculus is a formal system for computation, and it represents all first-order logic and naturally supports higher-order functions with constants, quantifiers, logical connectors, and lambda abstract.
FunQL is a variable-free language, and it encodes compositionality using nested function-argument structures.
SQL is the query language based upon relational algebra to handle relations among entities and variables in databases. 
The last two, ThingTalk Query Language~\cite{xu2020schema2qa} and Hierarchical intent and slot~\cite{gupta2018semantic} are recently proposed for Question Answering on Web and Task-Oriented Dialogue State Tracking, respectively. Python is a high-level, general-purpose programming language. Its design philosophy emphasizes code readability with the use of significant indentation.

\subsection*{A.4 \quad Dataset Statistics}
\begin{figure}[ht]
    \subfigure{
    
    \includegraphics[width=\linewidth]{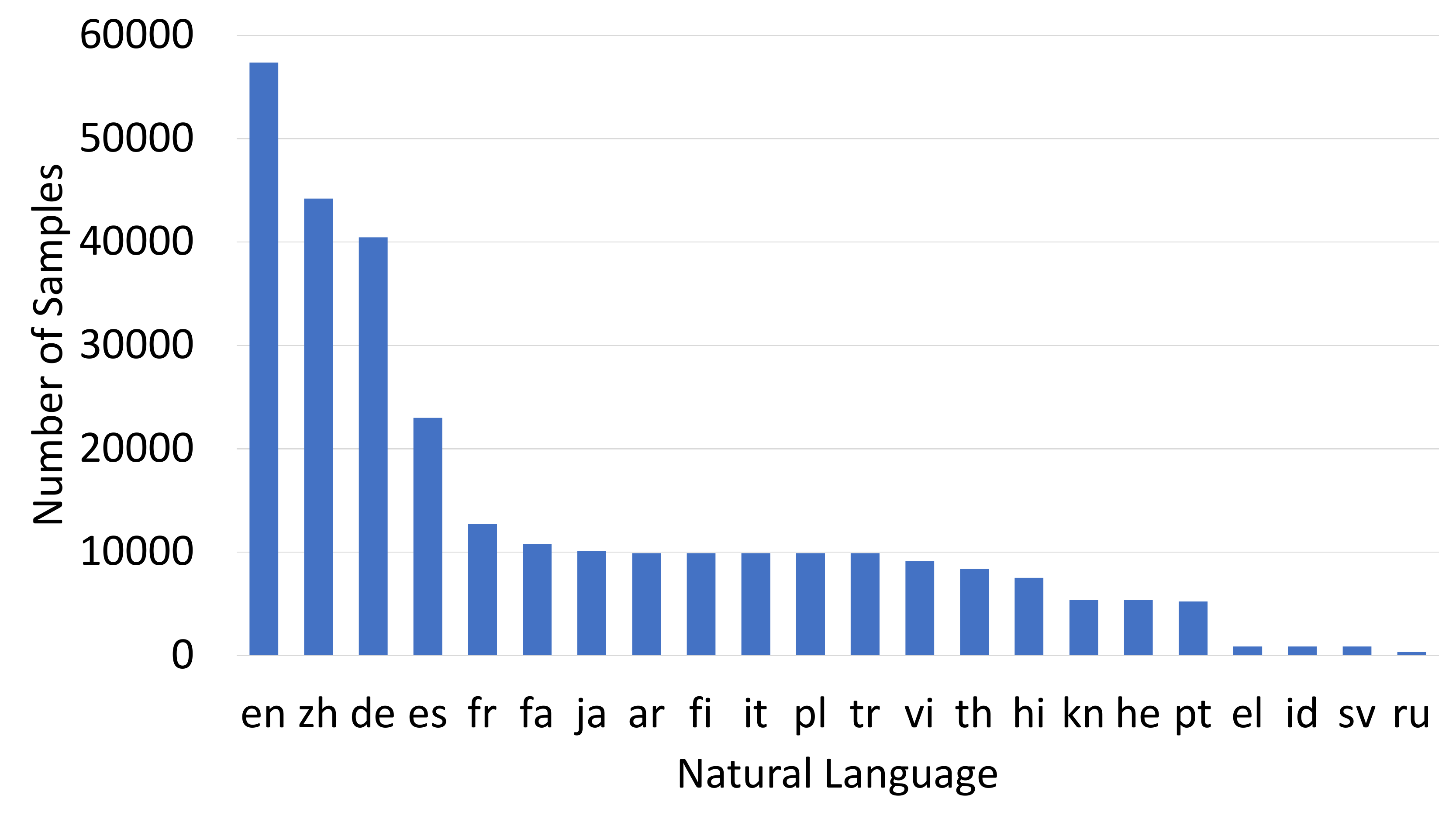} 

    }
    
    \subfigure{
    
    \includegraphics[width=\linewidth]{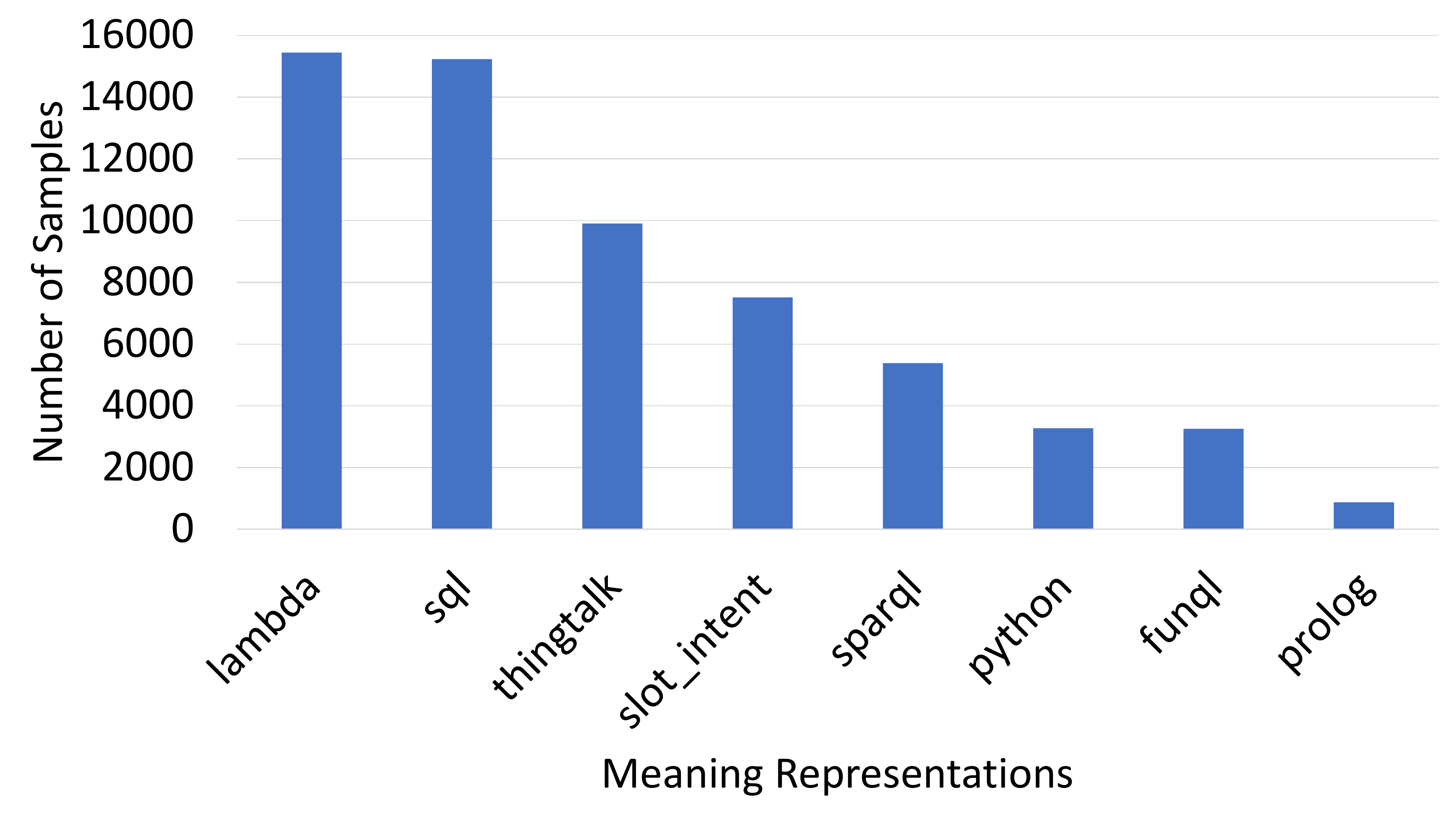}  
    }
    
    \caption{Distribution of 22 natural languages and 8 meaning representations. Each number of bar represents the sum of samples across all datasets.}
    \label{fig:distri}
\end{figure}

Figure~\ref{fig:distri} shows the statistics of the dataset. As can be seen, the top 3 NLs with the most samples in \ours are English, Chinese, and German, while the top 3 MRs are Lambda, SQL, and ThingTalk. 

\section{Experiment Details}
We introduce the training settings and input/output format for all experiments and settings in this section.

\subsection*{B.1 \quad Training Settings}
For experiments on LSTM model (Table~\ref{tab:lstm_mono}), we use OpenNMT\footnote{\url{https://opennmt.net/}} as the implementation. For Transformer-PTR models, we use Pytorch\footnote{\url{https://pytorch.org/}} as the implementation. For Codex and BLOOM models, we use OpenAI API\footnote{\url{https://platform.openai.com/docs/api-reference}} and Huggingface API\footnote{\url{https://huggingface.co/inference-api}}, respectively, and for mT5 and mBART models, we leverage Huggingface\footnote{\url{https://huggingface.co/}} as implementation. For each model, we train 300 epochs on MGeoquery due to the less number of training instances in this dataset and 100 epochs on the rest of the datasets. The learning rate is chosen from \{1e-5, 3e-5, 5e-5, 1e-4\} according to the parameter search on the dev set. 

For Codex and BLOOM, the maximum length of the generated sequence is set to 256 tokens. For Codex, the temperature is set to 0. For BLOOM, if the generated result does not contain complete MR, we append the generated results to the input and redo the generation and repeat this process over again until the generated result is complete. However, the maximum API call of one sample is set to 5 times. After 5 calls, we use the generated result as the final result. We use default settings for the rest of the parameters.

We run the model on 8 RTX A6000 GPUs, and it takes from hours to several days according to the data size. The model architecture from Huggingface is mT5-large, mBART-large, and mBERT-base. For Codex and BLOOM, we use code-davinci-002\footnote{code-davinci-002 has been deprecated}, and bigscience/bloom. The batch size is set to 16 for training mT5/mBART and 32 for training Transformer-PTR models. 

\subsection*{B.2 \quad Input/Output Format}
For input of the Transformer-PTR models, we directly feed the query into the model. For MSpider, we append the table to the end of the sequence with the format ``[CLS] Query [SEP] Schema name [SEP] Table 1 [SEP] Table 2 ...'', each table is represented by ``table name.column name''. We add ``table name.*'' to each table to represent all columns. For instance\footnote{In these examples, we use ``-'' to connect the words crossing lines.}:\\

\noindent{\ttfamily
[CLS] how many singers do we have? [SEP] * [SEP] stadium.* stadium.stadium\_id stadium.location stadium.name stadium.capacity stadium.highest stadium.lowest stadium.average [SEP] singer.* singer.singer\_id singer.name singer.country singer.song\_name singer.song\_release\_year singer.age singer.is\_male [SEP] concert.* concert.concert\_id concert.concert\_name concert.theme concert.stadium\_id concert.year [SEP] singer\_in\_concert.* singer\_in\_concert.concert\_id singer\_in\_concert.singer\_id [SEP]
}\\

As for the output, we scan the tokens in the label and replace the ones that appear in the source text with ``@ptrN'' where ``N'' is a natural number showing the index of the token in the source text. We remove the ``FROM'' clause in SQL. In this way, the pointer network can easily identify which tokens are copied from the source. For instance:\\

\noindent{\ttfamily
[CLS] select count ( @ptr19 ) [SEP] concert\_singer
}\\

For mT5 and mBART, we use the tokenizers provided by Huggingface to tokenize the queries. And for MSipder dataset, we append the table columns one by one to the tail, separated by ``||''. For instance: \\ 

\noindent{\ttfamily
how many singers do we have? || stadium.stadium\_id || stadium.location || stadium.name || stadium.capacity || stadium.highest || stadium.lowest || stadium.average || singer.singer\_id || singer.name || singer.country singer.song\_name || singer.song\_release\_year || singer.age singer.is\_male || concert.concert\_id || concert.concert\_name || concert.theme || concert.stadium\_id || concert.year || singer\_in\_concert.concert\_id || singer\_in\_concert.singer\_id
}\\

The output is simply the MR itself.\\

\noindent{\ttfamily
select count ( singer\_id ) from singer  
} \\

For Codex and BLOOM, we use 8-shot in-context learning~\citep{han2022folio}. Specifically, we concatenate 8 pairs of examples and a query as the input. For MSpider, we additionally list the information of the schema including table names and column names of each example. It is worth noting that the number of examples of BLOOM for in-context learning decrease to 4 on MATIS dataset and decreases to 1 on MSpider dataset because the number of tokens exceeds the input limit. The example of MSpider input is listed as follows:\\

\noindent{\ttfamily
\# Translate the following sentences into sql:\\
\\
\# Question:\\
\# Who performed the song named "Le Pop"?\\
\\
\# The information of tables:\\
\# 0. Table name is: Songs. The table columns are as follows: SongId, Title\\
\# 1. Table name is: Albums. The table columns are as follows: AId, Title, Year, Label, Type\\
\# 2. Table name is: Band. The table columns are \\
\\
      ---- 3 Tables Ignored ----\\
\\
\# 6. Table name is: Vocals. The table columns are as follows: SongId, Bandmate, Type\\
\\
\# Translation results are as follows:\\
\# SELECT T2.firstname ,  T2.lastname FROM Performance AS T1 JOIN Band AS T2 ON T1.bandmate  =  T2.id JOIN Songs AS T3 ON T3.SongId  =  T1.SongId WHERE T3.Title  =  "Le Pop"\\
\\
      ---- More Examples Ignored ---- \\
\\
\# Translate the following sentences into sql:\\
\\
\# Question:\\
\# Tell me the types of the policy used by the customer named "Dayana Robel".\\
\\
\# The information of tables:\\
\\
      ---- 6 Tables Ignored ---- \\
\\
\# Translation results are as follows:\\
}\\

The expected output is the MR with a starting symbol ``\#''.\\

\noindent{\ttfamily
\# SELECT DISTINCT t3.policy\_type\_code FROM customers AS t1 JOIN customers\_policies AS t2 ON t1.customer\_id  =  t2.customer\_id JOIN available\_policies AS t3 ON t2.policy\_id  =  t3.policy\_id WHERE t1.customer\_name  =  "Dayana Robel"
}\\

\subsection*{B.3 \quad  Experiment Path}
The experiments are done in the following order: we first evaluate 2 Enc-PTR and 2 Enc-Dec baseline models in the Monolingual setting. Then, we pick two of them with the best performance to evaluate on all the other settings. Finally, we evaluate LLMs using in-context learning in two finetuning-free settings.

\section{Results and Discussions}
This section lists the results for each NL and MR and introduces the comparison with SOTA, training data size and few-shot learning, and error analysis.

\subsection*{C.1 \quad Results for Each NL and MR}
We list some of the results of our models  on various datasets and languages. Table~\ref{tab:lstm_mono},~\ref{tab:mbert_mono},~\ref{tab:xlmr_mono},~\ref{tab:mt5_mono},~\ref{tab:mbart_mono} show the Monolingual performance of LSTM, mBERT+PTR, XLM-R+PTR, mBART, and mT5. Table~\ref{tab:xlm_monofew},~\ref{tab:mt5_monofew},~\ref{tab:codex_monofew},~\ref{tab:bloom_monofew} shows the Monolingual Few-Shot performance of XLM-R+PTR, mT5, Codex, and BLOOM.
Table~\ref{tab:xlmr_multi}, and~\ref{tab:mt5_multi} show the Multilingual performance of XLM-R+PTR, and mT5.
Table~\ref{tab:xlmr_crosszero},~\ref{tab:mt5_crosszero},~\ref{tab:codex_crosszero},~\ref{tab:bloom_crosszero} show the Cross-lingual Zero-Shot Transfer performance of XLM-R+PTR, mT5, Codex, and BLOOM.
Table~\ref{tab:xlmr_crossfew},~\ref{tab:mt5_crossfew} show the Cross-lingual Few-Shot Transfer performance of XLM-R+PTR, and mT5.

\subsection*{C.2 \quad Training Data Size and Few-shot Learning}

\begin{figure}[!t]
    \centering
    \includegraphics[width=\linewidth]{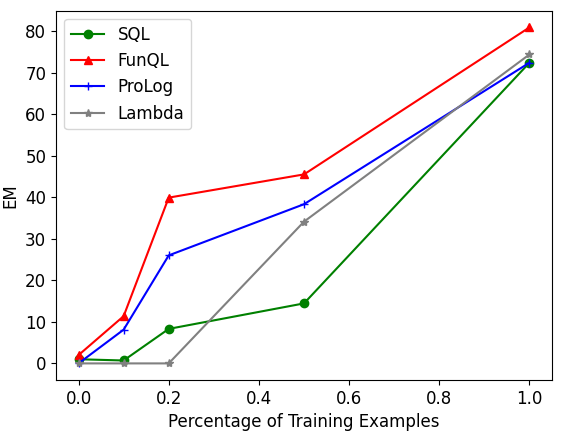}
    \caption{Exact Matching (EM) scores on the MGeoQuery dataset using mT5 as a monolingual learner.
    }
    \label{fig:few}
\end{figure}
\begin{table*}[!ht]
\centering
\begin{tabular}{@{}lll@{}}
\toprule
Error Type            & Description                                             & Proportion(\%) \\ \midrule
Syntax error          & Incorrect program syntax (invalid grammar)              & \textbf{17.14}          \\
Semantic error        &                                                         & \textbf{64.27}          \\
\;\;\;\;\;\;\;Token                 & Incorrect or missing column/value/operator              & \;\;\;\;\;22.85          \\
\;\;\;\;\;\;\;Structure             & Incorrect program structure (valid grammar)             & \;\;\;\;\;41.42          \\
Incorrect Exact Match & Incorrect exact match with the correct execution answer & \textbf{18.47}          \\ \bottomrule
\end{tabular}
\caption{Error analysis on MGeoQuery English test set. The MR is SQL.}
\label{tab:error}
\end{table*}
Figure~\ref{fig:few} displays the averaged Exact Matching scores (EM) across all languages on MGeoQuery dataset, where each line represents a meaning representation, and each dot on the line represents a few-shot experiment using such meaning representation. The X-axis is the percentage of data we use to train the model. Results show that the performance was largely influenced by the number of samples in the training set. The performance can be as high as 70\% if given sufficient data, while training on 10\% of training data may lead to 0 scores. Besides, among all four MRs, the performance of FunQL increases most steadily, showing its robustness.

\subsection*{C.3 \quad Error Analysis}
% \usepackage{booktabs}

We conduct error analysis on MGeoQuery dataset. First, we select the English split with SQL MR, and compare the golden MR and the predictions generated by mT5. We classify the errors into 4 types:
\begin{itemize}
    \item Syntax error: The prediction contains a syntax error. In other words, SQL engine can parse the predictions because of the grammar issues.
    \item Token error: one of the two types of semantic errors. Predictions contain wrong column names, values (such as strings and numbers), and operators (not including keywords). 
    \item Structure error: one of the two types of semantic errors. Predictions contain wrong structures. It means some keywords of SQL are incorrect or missing.
    \item Incorrect Exact Match: although the exact match shows the prediction is different from the golden one, the execution results are the same.
\end{itemize}
  As shown in Table~\ref{tab:error}, most of the errors are semantic errors (64.27\%) in which the structure error is around two times of token error (41.42\% vs. 22.85\%). Syntax error and incorrect exact match occupy around 18\% of errors respectively.

\newpage

\begin{table*}[ht!]
\centering
\resizebox{.9\textwidth}{!}{%
\begin{tabular}{@{}lllllllll@{}}
\toprule
                & MATIS & MGeoQuery & MSpider & MNLmaps & MOvernight & MCWQ$^\bigstar$ & MSchema2QA & MTOP \\ \midrule
English         & 48.9 & 76.8 & 15.8 & 72.2 & 22.4 & 92 & 48.1 & 78.6 \\
Arabic          & --   & -- & -- & -- & -- & -- & 33.1 & -- \\
Chinese         & 44.6 & 61.2 & 10.2 & -- & 20.8 & 75.1 & 35.9 & -- \\
Farsi           & -- & 52.0 & -- & -- & -- & -- & 24.4 & -- \\
Finnish         & -- & -- & -- & -- &  -- & -- & 24.7 & -- \\
French          & 47.5 & -- & -- & --  & -- & -- & -- & 60.8 \\
German          & 47.7 & 59.5 & -- & 64.9 & 2.1 & -- & 38.3 & 58.5 \\
Greek           & -- & 51.4 & -- & --  & -- & -- & -- & -- \\
Hebrew          & -- & -- & -- & --  & -- & 74.0 & -- & -- \\
Hindi           & -- & -- & -- & --  & -- & -- & -- & 58.6 \\
Indonesian      & -- & 69.3 & --  & -- & -- & -- & -- & -- \\
Italian         & -- & -- & --  & -- & -- & -- & 33.8 & -- \\
Japanese        & 2.7 & -- & -- & -- & -- & -- & 49.6 & -- \\
Kannada         & -- & -- & -- & -- & -- & 77.7 & -- & -- \\
Polish          & -- & -- & -- & -- & -- & -- & 31.4 & -- \\
Portuguese      & 46.4 & -- & -- & -- & -- & -- & -- & -- \\
Spanish         & 7.2 & -- & -- & -- & -- & -- & 44.5 & 63.8 \\
Swedish         & -- & 63.3 & -- & -- & -- & -- & -- & -- \\
Thai            & -- & 48.6 & -- & -- & -- & -- & -- & 60.0 \\
Turkish         & -- & -- & -- & -- & -- & -- & 41.4 & -- \\
Vietnamese      & -- & -- & 8.6 & -- & -- & -- & -- & -- \\
\bottomrule
\end{tabular}
}
\caption{The performance of LSTM with Monolingual setting on different datasets and different languages.$^\bigstar$ We use random split in LSTM experiments rather than MCD3 split for MCWQ dataset.}
\label{tab:lstm_mono}
\end{table*}

\begin{table*}[ht!]
\centering
\resizebox{\textwidth}{!}{%
\begin{tabular}{@{}llllllllll@{}}
\toprule
                & MATIS & MGeoQuery & MSpider & MNLmaps & MOvernight & MCWQ & MSchema2QA & MTOP & MCoNaLa\\ \midrule
English         & 37.33 & 88.08 & 55.4 & 85.8 & 61.82 & 35.49 & 64.98 & 86.58 & 5.87\\
Arabic          & --   & -- & -- & -- & -- & -- & 48.09 & -- & --\\
Chinese         & 32.26 & 63.9 & 42.6 & -- & 53.36 & 22.38 & 43.87 & -- & --\\
Farsi           & -- & 80.86 & -- & -- & -- & -- & 46.65 & -- & --\\
Finnish         & -- & -- & -- & -- &  -- & -- & 50.26 & -- & --\\
French          & 34.21 & -- & -- & --  & -- & -- & -- & 75.18 & --\\
German          & 37.56 & 85.92 & -- & 81.84 & 57.22 & -- & 60.56 & 73.16 & --\\
Greek           & -- & 86.64 & -- & --  & -- & -- & -- & -- & --\\
Hebrew          & -- & -- & -- & --  & -- & 24.38 & -- & -- & --\\
Hindi           & -- & -- & -- & --  & -- & -- & -- & 70.97 & --\\
Indonesian      & -- & 84.84 & --  & -- & -- & -- & -- & -- & --\\
Italian         & -- & -- & --  & -- & -- & -- & 50.26 & -- & --\\
Japanese        & -- & -- & -- & -- & -- & -- & 48.97 & -- & --\\
Kannada         & -- & -- & -- & -- & -- & 11.57 & -- & -- & --\\
Polish          & -- & -- & -- & -- & -- & -- & 45.31 & -- & --\\
Portuguese      & 36.64 & -- & -- & -- & -- & -- & -- & -- & --\\
Russian         & -- & -- & -- & -- &  -- & -- & -- & -- & --\\
Spanish         & 5.76 & -- & -- & -- & -- & -- & 62.51 & 77.2 & --\\
Swedish         & -- & 87.36 & -- & -- & -- & -- & -- & -- & --\\
Thai            & -- & 81.58 & -- & -- & -- & -- & -- & 69.36 & --\\
Turkish         & -- & -- & -- & -- & -- & -- & 56.33 & -- & --\\
Vietnamese      & -- & -- & 23.2 & -- & -- & -- & -- & -- & --\\
\bottomrule
\end{tabular}
}
\caption{The performance of mBERT+PTR with Monolingual setting on different datasets and different languages.}
\label{tab:mbert_mono}
\end{table*}

\begin{table*}[ht!]
\centering
\resizebox{\textwidth}{!}{
\begin{tabular}{@{}llllllllll@{}}
\toprule
                & MATIS & MGeoQuery & MSpider & MNLmaps & MOvernight & MCWQ & MSchema2QA & MTOP & MCoNaLa \\ \midrule
English         & 36.71 & 88.45 & 53.1 & 86.02 & 62.99 & 37.19 & 73.53 & 90.54 & 7.69\\
Arabic          & --   & -- & -- & -- & -- & -- & 58.08 & -- & --\\
Chinese         & 34.91 & 77.98 & 44.1 & -- & 56.93 & 19.29 & 48.92 & -- & --\\
Farsi           & -- & 81.23 & -- & -- & -- & -- & 60.56 & -- & --\\
Finnish         & -- & -- & -- & -- &  -- & -- & 64.26 & -- & --\\
French          & 38.31 & -- & -- & --  & -- & -- & -- & 78.58 & --\\
German          & 38.28 & 89.17 & -- & 84.32 & 59.27 & -- & 68.59 & 79.22 & --\\
Greek           & -- & 85.92 & -- & --  & -- & -- & -- & -- & --\\
Hebrew          & -- & -- & -- & --  & -- & 14.66 & -- & -- & --\\
Hindi           & -- & -- & -- & --  & -- & -- & -- & 77.93 & --\\
Indonesian      & -- & 88.81 & --  & -- & -- & -- & -- & -- & --\\
Italian         & -- & -- & --  & -- & -- & -- & 63.44 & -- & --\\
Japanese        & -- & -- & -- & -- & -- & -- & 55.26 & -- & --\\
Kannada         & -- & -- & -- & -- & -- & 22.99& -- & -- & --\\
Polish          & -- & -- & -- & -- & -- & -- & 55.82 & -- & --\\
Portuguese      & 34.01 & -- & -- & -- & -- & -- & -- & -- & --\\
Russian         & -- & -- & -- & -- &  -- & -- & -- & -- & -- \\
Spanish         & 5.63 & -- & -- & -- & -- & -- & 68.59 & 81.16 & --\\
Swedish         & -- & 89.17 & -- & -- & -- & -- & -- & -- & --\\
Thai            & -- & 85.56 & -- & -- & -- & -- & -- & 74.7 & --\\
Turkish         & -- & -- & -- & -- & -- & -- & 69 & -- & --\\
Vietnamese      & -- & -- & 44.7 & -- & -- & -- & -- & -- & --\\
\bottomrule
\end{tabular}
}
\caption{The performance of XLM-R+PTR with Monolingual setting on different datasets and different languages.}
\label{tab:xlmr_mono}
\end{table*}

\begin{table*}[ht!]
\centering
\resizebox{\textwidth}{!}{
\begin{tabular}{@{}llllllllll@{}}
\toprule
                & MATIS & MGeoQuery & MSpider & MNLmaps & MOvernight & MCWQ & MSchema2QA & MTOP & MCoNaLa \\ \midrule
English         & 45.72 & 74.01 & 52.32 & 86.82 & 65.18 & 38.12 & 57.16 & 87.87 & 6.78\\
Arabic          & --   & -- & -- & -- & -- & -- & 44.59 & -- & --\\
Chinese         & 45.72 & 65.34 & 16.48 & -- & 56.93 & 25.35 & 42.95 & -- & --\\
Farsi           & -- & 59.57 & -- & -- & -- & -- & 38.72 & -- & --\\
Finnish         & -- & -- & -- & -- &  -- & -- & 54.48 & -- & --\\
French          & 47.97 & -- & -- & --  & -- & -- & -- & 74.05 & --\\
German          & 50.23 & 57.76 & -- & 79.55 & 56.68 & -- & 59.11 & 75.18 & --\\
Greek           & -- & 49.46 & -- & --  & -- & -- & -- & -- & --\\
Hebrew          & -- & -- & -- & --  & -- & 33.95 & -- & -- & --\\
Hindi           & -- & -- & -- & --  & -- & -- & -- & 72.59 & --\\
Indonesian      & -- & 74.01 & --  & -- & -- & -- & -- & -- & --\\
Italian         & -- & -- & --  & -- & -- & -- & 46.65 & -- & --\\
Japanese        & -- & -- & -- & -- & -- & -- & 53.76 & -- & --\\
Kannada         & -- & -- & -- & -- & -- & 22.69 & -- & -- & --\\
Polish          & -- & -- & -- & -- & -- & -- & 43.56 & -- & --\\
Portuguese      & 43.47 & -- & -- & -- & -- & -- & -- & -- & --\\
Russian         & -- & -- & -- & -- &  -- & -- & -- & -- & -- \\
Spanish         & 18.47 & -- & -- & -- & -- & -- & 57.67 & 78.66 & --\\
Swedish         & -- & 68.95 & -- & -- & -- & -- & -- & -- & --\\
Thai            & -- & 58.12 & -- & -- & -- & -- & -- & 66.21 & --\\
Turkish         & -- & -- & -- & -- & -- & -- & 46.65 & -- & --\\
Vietnamese      & -- & -- & 14.31 & -- & -- & -- & -- & -- & --\\
\bottomrule
\end{tabular}
}
\caption{The performance of mBART with Monolingual setting on different datasets and different languages.}
\label{tab:mbart_mono}
\end{table*}

\begin{table*}[ht!]
\centering
\resizebox{\textwidth}{!}{
\begin{tabular}{@{}llllllllll@{}}
\toprule
                & MATIS & MGeoQuery & MSpider & MNLmaps & MOvernight & MCWQ & MSchema2QA & MTOP & MCoNaLa \\ \midrule
English         & 53.60 & 89.89 & 68.30 & 92.73 & 69.38 & 39.29 & 76.00 & 91.67 & 10.29\\
Arabic          & --    & --    & --    & --    & --    & --    & 53.55 & --    & --\\
Chinese         & 52.48 & 77.62 & 54.90 & --    & 62.59 & 24.56 & 56.54 & --    & --\\
Farsi           & --    & 75.45 & --    & --    & --    & --    & 60.25 & --    & --\\
Finnish         & --    & --    & --    & --    &  --   & --    & 68.28 & --    & --\\
French          & 53.60 & --    & --    & --    & --    & --    & --    & 82.30 & --\\
German          & 52.93 & 71.83 & --    & 90.57 & 66.90 & --    & 72.19 & 82.38 & --\\
Greek           & --    & 85.92 & --    & --    & --    & --    & --    & --    & --\\
Hebrew          & --    & --    & --    & --    & --    & 33.02 & --    & --    & --\\
Hindi           & --    & --    & --    & --    & --    & --    & --    & 78.98 & --\\
Indonesian      & --    & 87.00 & --    & --    & --    & --    & --    & --    & --\\
Italian         & --    & --    & --    & --    & --    & --    & 67.97 & --    & --\\
Japanese        & --    & --    & --    & --    & --    & --    & 62.41 & --    & --\\
Kannada         & --    & --    & --    & --    & --    & 23.74 & --    & --    & --\\
Polish          & --    & --    & --    & --    & --    & --    & 60.87 & --    & --\\
Portuguese      & 53.15 & --    & --    & --    & --    & --    & --    & --    & --\\
Russian         & --    & --    & --    & --    & --    & --    & --    & --    & --\\
Spanish         & 53.13 & --    & --    & --    & --    & --    & 68.69 & 83.91 & --\\
Swedish         & --    & 84.48 & --    & --    & --    & --    & --    & --    & --\\
Thai            & --    & 76.17 & --    & --    & --    & --    & --    & 71.71 & --\\
Turkish         & --    & --    & --    & --    & --    & --    & 70.03 & --    & --\\
Vietnamese      & --    & --    & 57.15 & --    & --    & --    & --    & --    & --\\
\bottomrule
\end{tabular}
}
\caption{The performance of mT5 with Monolingual setting on different datasets and different languages.}
\label{tab:mt5_mono}
\end{table*}

\begin{table*}[ht!]
\centering
\resizebox{\textwidth}{!}{
\begin{tabular}{@{}llllllllll@{}}
\toprule
                & MATIS & MGeoQuery & MSpider & MNLmaps & MOvernight & MCWQ & MSchema2QA & MTOP & MCoNaLa \\ \midrule
English         & 29.50 & 27.01 & 43.44 & 20.68 & 47.88 &  9.41 & 58.91 & 69.36 & 0.38 \\
Arabic          & --    & --    & --    & --    & --    & --    & 48.71 & --    & -- \\
Chinese         & 28.11 &  6.51 & 33.76 & --    & 34.85 &  6.02 & 34.91 & --    & -- \\
Farsi           & --    &  6.04 & --    & --    & --    & --    & 37.69 & --    & -- \\
Finnish         & --    & --    & --    & --    &  --   & --    & 56.13 & --    & -- \\
French          & 37.80 & --    & --    & --    & --    & --    & --    & 58.21 & -- \\
German          &  5.85 & 21.50 & --    & 18.86 & 39.49 & --    & 57.57 & 60.55 & -- \\
Greek           & --    & 26.20 & --    & --    & --    & --    & --    & --    & -- \\
Hebrew          & --    & --    & --    & --    & --    &  1.08 & --    & --    & -- \\
Hindi           & --    & --    & --    & --    & --    & --    & --    & 59.66 & -- \\
Indonesian      & --    & 25.47 & --    & --    & --    & --    & --    & --    & -- \\
Italian         & --    & --    & --    & --    & --    & --    & 48.09 & --    & -- \\
Japanese        & --    & --    & --    & --    & --    & --    & 41.55 & --    & -- \\
Kannada         & --    & --    & --    & --    & --    &  6.02 & --    & --    & -- \\
Polish          & --    & --    & --    & --    & --    & --    & 40.99 & --    & -- \\
Portuguese      & 37.33 & --    & --    & --    & --    & --    & --    & --    & -- \\
Russian         & --    & --    & --    & --    & --    & --    & --    & --    & -- \\
Spanish         &  2.02 & --    & --    & --    & --    & --    & 54.48 & 62.09 & -- \\
Swedish         & --    & 23.40 & --    & --    & --    & --    & --    & --    & -- \\
Thai            & --    &  7.14 & --    & --    & --    & --    & --    & 52.63 & -- \\
Turkish         & --    & --    & --    & --    & --    & --    & 59.94 & --    & -- \\
Vietnamese      & --    & --    & 30.92 & --    & --    & --    & --    & --    & -- \\

\bottomrule
\end{tabular}
}
\caption{The performance of XLM-R+PTR with Monolingual Few-shot setting on different datasets and different languages.}
\label{tab:xlm_monofew}
\end{table*}

\begin{table*}[ht!]
\centering
\resizebox{\textwidth}{!}{
\begin{tabular}{@{}llllllllll@{}}
\toprule
                & MATIS & MGeoQuery & MSpider & MNLmaps & MOvernight & MCWQ & MSchema2QA & MTOP & MCoNaLa \\ \midrule
English         & 31.98 & 33.26 & 40.81 & 36.25 & 59.48 & 10.80 & 39.24 & 72.43 & 1.05 \\
Arabic          & --    & --    & --    & --    & --    & --    & 24.20 & --    & -- \\
Chinese         & 32.88 & 16.25 & 33.46 & --    & 48.47 &  4.63 & 19.26 & --    & -- \\
Farsi           & --    & 17.69 & --    & --    & --    & --    & 23.27 & --    & -- \\
Finnish         & --    & --    & --    & --    &  --   & --    & 35.84 & --    & -- \\
French          & 28.60 & --    & --    & --    & --    & --    & --    & 62.81 & -- \\
German          & 20.27 & 23.82 & --    & 26.93 & 52.81 & --    & 36.05 & 60.91 & -- \\
Greek           & --    & 29.88 & --    & --    & --    & --    & --    & --    & -- \\
Hebrew          & --    & --    & --    & --    & --    &  6.20 & --    & --    & -- \\
Hindi           & --    & --    & --    & --    & --    & --    & --    & 61.20 & -- \\
Indonesian      & --    & 30.42 & --    & --    & --    & --    & --    & --    & -- \\
Italian         & --    & --    & --    & --    & --    & --    & 45.73 & --    & -- \\
Japanese        & --    & --    & --    & --    & --    & --    & 29.66 & --    & -- \\
Kannada         & --    & --    & --    & --    & --    &  9.10 & --    & --    & -- \\
Polish          & --    & --    & --    & --    & --    & --    & 28.94 & --    & -- \\
Portuguese      & 27.93 & --    & --    & --    & --    & --    & --    & --    & -- \\
Russian         & --    & --    & --    & --    & --    & --    & --    & --    & -- \\
Spanish         &  7.43 & --    & --    & --    & --    & --    & 47.89 & 59.90 & -- \\
Swedish         & --    & 32.40 & --    & --    & --    & --    & --    & --    & -- \\
Thai            & --    & 21.21 & --    & --    & --    & --    & --    & 54.16 & -- \\
Turkish         & --    & --    & --    & --    & --    & --    & 35.84 & --    & -- \\
Vietnamese      & --    & --    & 37.04 & --    & --    & --    & --    & --    & -- \\
\bottomrule
\end{tabular}
}
\caption{The performance of mT5 with Monolingual Few-shot setting on different datasets and different languages.}
\label{tab:mt5_monofew}
\end{table*}

\begin{table*}[ht!]
\centering
\resizebox{\textwidth}{!}{
\begin{tabular}{@{}llllllllll@{}}
\toprule
            & MATIS & MGeoQuery & MSpider & MNLmaps & MOvernight & MCWQ & MSchema2QA & MTOP & MCoNaLa \\\midrule
English     & 17.79   & 34.39   & 34.43   & 36.82   & 4.34    & 4.48    & 22.97   & 20.21 & 13.87     \\
Arabic      & --      & --      & --      & --      & --      & --      & 16.79   & --    & --     \\
Chinese     & 16.89   & 31.77   & 27.85   & --      & 2.74    & 3.86    & 18.85   & --    & --     \\
Farsi       & --      & 27.71   & --      & --      & --      & --      & 17.61   & --    & --     \\
Finnish     & --      & --      & --      & --      & --      & --      & 21.52   & --    & --     \\
French      & 18.47   & --      & --      & --      & --      & --      & --      & 17.46 & --     \\
German      & 18.24   & 31.59   & --      & 31.70   & 3.21    & --      & 20.60   & 18.51 & --     \\
Greek       & --      & 33.03   & --      & --      & --      & --      & --      & --    & --     \\
Hebrew      & --      & --      & --      & --      & --      & 2.47    & --      & --    & --     \\
Hindi       & --      & --      & --      & --      & --      & --      & --      & 0.49  & --     \\
Indonesian  & --      & 32.49   & --      & --      & --      & --      & --      & --    & --     \\
Italian     & --      & --      & --      & --      & --      & --      & 24.30   & --    & --     \\
Japanese    & --      & --      & --      & --      & --      & --      & 19.36   & --    & --     \\
Kannada     & --      & --      & --      & --      & --      & 0.93    & --      & --    & --     \\
Polish      & --      & --      & --      & --      & --      & --      & 20.70   & --    & --     \\
Portuguese  & 18.24   & --      & --      & --      & --      & --      & --      & --    & --     \\
Russian     & --      & --      & --      & --      & --      & --      & --      & --    & -- \\
Spanish     & 18.47   & --      & --      & --      & --      & --      & 24.30   & 1.13  & --     \\
Swedish     & --      & 33.85   & --      & --      & --      & --      & --      & --    & --     \\
Thai        & --      & 30.60   & --      & --      & --      & --      & --      & 2.67  & --     \\
Turkish     & --      & --      & --      & --      & --      & --      & 30.79   & --    & --     \\
Vietnamese  & --      & --      & 29.69   & --      & --      & --      & --      & --    & --     \\
\bottomrule

\end{tabular}
}
\caption{The performance of Codex with Monolingual Few-shot setting on different datasets and different languages.}
\label{tab:codex_monofew}
\end{table*}

\begin{table*}[ht!]
\centering
\resizebox{\textwidth}{!}{
\begin{tabular}{@{}llllllllll@{}}
\toprule
            & MATIS & MGeoQuery & MSpider & MNLmaps & MOvernight & MCWQ & MSchema2QA & MTOP & MCoNaLa \\\midrule
English     & 0.00    & 21.66   & 2.22    & 15.23   & 0.91    & 0.00    & 9.68    & 7.03    & 8.40    \\
Arabic      & --      & --      & --      & --      & --      & --      & 5.87    & --      & --      \\
Chinese     & 0.00    & 20.76   & 2.71    & --      & 0.62    & 0.00    & 4.43    & --      & --      \\
Farsi       & --      & 11.64   & --      & --      & --      & --      & 1.96    & --      & --      \\
Finnish     & --      & --      & --      & --      & --      & --      & 3.71    & --      & --      \\
French      & 0.00    & --      & --      & --      & --      & --      & --      & 5.25    & --      \\
German      & 0.00    & 19.86   & --      & 9.09    & 0.33    & --      & 8.24    & 5.66    & --      \\
Greek       & --      & 18.05   & --      & --      & --      & --      & --      & --      & --      \\
Hebrew      & --      & --      & --      & --      & --      & 0.00    & --      & --      & --      \\
Hindi       & --      & --      & --      & --      & --      & --      & --      & 5.50    & --      \\
Indonesian  & --      & 22.48   & --      & --      & --      & --      & --      & --      & --      \\
Italian     & --      & --      & --      & --      & --      & --      & 5.77    & --      & --      \\
Japanese    & --      & --      & --      & --      & --      & --      & 4.02    & --      & --      \\
Kannada     & --      & --      & --      & --      & --      & 0.00    & --      & --      & --      \\
Polish      & --      & --      & --      & --      & --      & --      & 2.99    & --      & --      \\
Portuguese  & 0.00    & --      & --      & --      & --      & --      & --      & --      & --      \\
Russian     & --      & --      & --      & --      & --      & --      & --      & --    & -- \\
Spanish     & 0.00    & --      & --      & --      & --      & --      & 8.75    & 4.77    & --      \\
Swedish     & --      & 19.59   & --      & --      & --      & --      & --      & --      & --      \\
Thai        & --      & 8.66    & --      & --      & --      & --      & --      & 2.75    & --      \\
Turkish     & --      & --      & --      & --      & --      & --      & 1.96    & --      & --      \\
Vietnamese  & --      & --      & 1.45    & --      & --      & --      & --      & --      & --      \\
\bottomrule

\end{tabular}
}
\caption{The performance of BLOOM with Monolingual Few-shot setting on different datasets and different languages.}
\label{tab:bloom_monofew}
\end{table*}

\begin{table*}[ht!]
\centering
\resizebox{.9\textwidth}{!}{
\begin{tabular}{@{}lllllllll@{}}
\toprule
            & MATIS & MGeoQuery & MSpider & MNLmaps & MOvernight & MCWQ & MSchema2QA & MTOP \\\midrule
English     & 40.05   & 76.42   & 36.63   & 85.91   & 63.69   & 32.72   & 61.32   & 89.57   \\
Arabic      & --      & --      & --      & --      & --      & --      & 65.19   & --      \\
Chinese     & 40.84   & 69.37   & 45.70   & --      & 58.07   & 31.94   & 68.25   & --      \\
Farsi       & --      & 66.85   & --      & --      & --      & --      & 62.62   & --      \\
Finnish     & --      & --      & --      & --      & --      & --      & 62.00   & --      \\
French      & 41.30   & --      & --      & --      & --      & --      & --      & 82.54   \\
German      & 39.68   & 68.38   & --      & 85.91   & 61.35   & --      & 70.44   & 81.00   \\
Greek       & --      & 73.80   & --      & --      & --      & --      & --      & --      \\
Hebrew      & --      & --      & --      & --      & --      & 28.86   & --      & --      \\
Hindi       & --      & --      & --      & --      & --      & --      & --      & 78.74   \\
Indonesian  & --      & 75.24   & --      & --      & --      & --      & --      & --      \\
Italian     & --      & --      & --      & --      & --      & --      & 57.88   & --      \\
Japanese    & --      & --      & --      & --      & --      & --      & 59.32   & --      \\
Kannada     & --      & --      & --      & --      & --      & 29.63   & --      & --      \\
Polish      & --      & --      & --      & --      & --      & --      & 64.12   & --      \\
Portuguese  & 42.46   & --      & --      & --      & --      & --      & --      & --      \\
Spanish     & 34.03   & --      & --      & --      & --      & --      & 54.58   & 81.73   \\
Swedish     & --      & 74.52   & --      & --      & --      & --      & --      & --      \\
Thai        & --      & 66.22   & --      & --      & --      & --      & --      & 76.48   \\
Turkish     & --      & --      & --      & --      & --      & --      & 54.27   & --      \\
Vietnamese  & --      & --      & 38.28   & --      & --      & --      & --      & --      \\
\bottomrule

\end{tabular}
}
\caption{The performance of XLM-R+PTR with Multilingual setting on different datasets and different languages.}
\label{tab:xlmr_multi}
\end{table*}

\begin{table*}[ht!]
\centering
\resizebox{.9\textwidth}{!}{
\begin{tabular}{@{}lllllllll@{}}
\toprule
            & MATIS & MGeoQuery & MSpider & MNLmaps & MOvernight & MCWQ & MSchema2QA & MTOP \\\midrule
English     & 58.45   & 82.04   & 36.07   & 92.27   & 70.33   & 29.94   & 69.52   & 90.61   \\
Arabic      & --      & --      & --      & --      & --      & --      & 56.09   & --      \\
Chinese     & 49.83   & 75.99   & 30.66   & --      & 63.98   & 28.24   & 58.15   & --      \\
Farsi       & --      & 71.48   & --      & --      & --      & --      & 55.17   & --      \\
Finnish     & --      & --      & --      & --      & --      & --      & 62.96   & --      \\
French      & 55.00   & --      & --      & --      & --      & --      & --      & 78.47   \\
German      & 60.12   & 74.19   & --      & 90.34   & 68.36   & --      & 65.27   & 83.46   \\
Greek       & --      & 79.61   & --      & --      & --      & --      & --      & --      \\
Hebrew      & --      & --      & --      & --      & --      & 28.55   & --      & --      \\
Hindi       & --      & --      & --      & --      & --      & --      & --      & 85.58   \\
Indonesian  & --      & 80.42   & --      & --      & --      & --      & --      & --      \\
Italian     & --      & --      & --      & --      & --      & --      & 58.10   & --      \\
Japanese    & --      & --      & --      & --      & --      & --      & 62.55   & --      \\
Kannada     & --      & --      & --      & --      & --      & 27.31   & --      & --      \\
Polish      & --      & --      & --      & --      & --      & --      & 56.23   & --      \\
Portuguese  & 48.47   & --      & --      & --      & --      & --      & --      & --      \\
Spanish     & 54.85   & --      & --      & --      & --      & --      & 63.31   & 84.12   \\
Swedish     & --      & 79.33   & --      & --      & --      & --      & --      & --      \\
Thai        & --      & 69.50   & --      & --      & --      & --      & --      & 75.48   \\
Turkish     & --      & --      & --      & --      & --      & --      & 62.78   & --      \\
Vietnamese  & --      & --      & 30.17   & --      & --      & --      & --      & --      \\
\bottomrule
\end{tabular}
}
\caption{The performance of mT5 with Multilingual setting on different datasets and different languages.}
\label{tab:mt5_multi}
\end{table*}

\begin{table*}[ht!]
\centering
\resizebox{\textwidth}{!}{
\begin{tabular}{@{}lllllllllll@{}}
\toprule
            & MATIS & MGeoQuery & MSpider & MNLmaps & MOvernight & MCWQ-MCD3 & MSchema2QA & MTOP  & MCoNaLa \\\midrule
Arabic      & --      & --      & --      & --      & --      & --      & 3.91    & --    & --  \\
Chinese     & 0.92    & 12.83   & 20.30   & --      & 23.80   & 2.16    & 0.51    & --    & --  \\
Farsi       & --      & 17.80   & --      & --      & --      & --      & 18.33   & --    & --  \\
Finnish     & --      & --      & --      & --      & --      & --      & 26.98   & --    & --  \\
French      & 2.15    & --      & --      & --      & --      & --      & --      & 59.90 & --  \\
German      & 1.61    & 51.13   & --      & 60.23   & 49.74   & --      & 40.37   & 56.27 & --  \\
Greek       & --      & 58.44   & --      & --      & --      & --      & --      & --    & --  \\
Hebrew      & --      & --      & --      & --      & --      & 5.56    & --      & --    & --  \\
Hindi       & --      & --      & --      & --      & --      & --      & --      & 44.14 & --  \\
Indonesian  & --      & 56.19   & --      & --      & --      & --      & --      & --    & --  \\
Italian     & --      & --      & --      & --      & --      & --      & 32.96   & --    & --  \\
Japanese    & --      & --      & --      & --      & --      & --      & 0.31    & --    & 0.20  \\
Kannada     & --      & --      & --      & --      & --      & 5.09    & --      & --    & --  \\
Polish      & --      & --      & --      & --      & --      & --      & 29.97   & --    & --  \\
Portuguese  & 0.23    & --      & --      & --      & --      & --      & --      & --    & --  \\
Russian     & --      & --      & --      & --      & --      & --      & --      & --    & 0.07 \\
Spanish     & 25.35   & --      & --      & --      & --      & --      & 39.24   & 62.65 & 0.10  \\
Swedish     & --      & 65.22   & --      & --      & --      & --      & --      & --    & --  \\
Thai        & --      & 17.35   & --      & --      & --      & --      & --      & 34.36 & --  \\
Turkish     & --      & --      & --      & --      & --      & --      & 9.58    & --    & --  \\
Vietnamese  & --      & --      & 16.76   & --      & --      & --      & --      & --    & --  \\
\bottomrule
\end{tabular}
}
\caption{The performance of XLM-R+PTR with Cross-lingual Zero-Shot Transfer setting on different datasets and different languages.}
\label{tab:xlmr_crosszero}
\end{table*}

\begin{table*}[ht!]
\centering
\resizebox{\textwidth}{!}{
\begin{tabular}{@{}llllllllll@{}}
\toprule
                & MATIS & MGeoQuery & MSpider & MNLmaps & MOvernight & MCWQ & MSchema2QA & MTOP & MCoNaLa \\ \midrule
Arabic          & --    & --    & --    & --    & --    & --   & 38.31 & -- & --\\
Chinese         & 18.02 & 17.69 & 38.59 & --    & 45.91 & 1.39 & 26.67 & -- & --\\
Farsi           & --    & 25.27 & --    & --    & --    & --   & 41.40 & -- & --\\
Finnish         & --    & --    & --    & --    & --    & --   & 50.26 & -- & --\\
French          & 33.56 & --    & --    & --    & --    & --   & --    & 61.92 & --\\
German          & 34.68 & 53.43 & --    & 34.89 & 59.45 & --   & 59.32 & 52.22 & --\\
Greek           & --    & 50.90 & --    & --    & --    & --   & --    & -- & --\\
Hebrew          & --    & --    & --    & --    & --    & 5.86 & --    & -- & --\\
Hindi           & --    & --    & --    & --    & --    & --   & --    & 35.89 & --\\
Indonesian      & --    & 42.24 & --    & --    & --    & --   & --    & -- & --\\
Italian         & --    & --    & --    & --    & --    & --   & 58.50 & -- & --\\
Japanese        & --    & --    & --    & --    & --    & --   & 11.64 & -- & 1.43\\
Kannada         & --    & --    & --    & --    & --    & 4.94 & --    & -- & --\\
Polish          & --    & --    & --    & --    & --    & --   & 49.95 & -- & --\\
Portuguese      & 34.46 & --    & --    & --    & --    & --   & --    & -- & --\\
Russian         & --    & --    & --    & --    & --    & --   & --    & -- & 0.29\\
Spanish         & 38.51 & --    & --    & --    & --    & --   & 55.82 & 61.36 & 0.59\\
Swedish         & --    & 68.23 & --    & --    & --    & --   & --    & -- & --\\
Thai            & --    & 18.05 & --    & --    & --    & --   & --    & 39.53 & --\\
Turkish         & --    & --    & --    & --    & --    & --   & 48.51 & -- & --\\
Vietnamese      & --    & --    & 45.26 & --    & --    & --   & --    & -- & --\\
\bottomrule      
\end{tabular}
}
\caption{The performance of mT5 with Cross-lingual Zero-Shot Transfer setting on different datasets and different languages.}
\label{tab:mt5_crosszero}
\end{table*}

\begin{table*}[ht!]
\centering
\resizebox{\textwidth}{!}{
\begin{tabular}{@{}llllllllll@{}}
\toprule
            & MATIS & MGeoQuery & MSpider & MNLmaps & MOvernight & MCWQ & MSchema2QA & MTOP & MCoNaLa \\\midrule
Arabic      & --      & --      & --      & --      & --      & --      & 17.82   & --      & --      \\
Chinese     & 12.61   & 26.62   & 27.18   & --      & 2.70    & 3.55    & 17.40   & --      & --      \\
Farsi       & --      & 25.36   & --      & --      & --      & --      & 16.79   & --      & --      \\
Finnish     & --      & --      & --      & --      & --      & --      & 22.35   & --      & --      \\
French      & 17.57   & --      & --      & --      & --      & --      & --      & 15.76   & --      \\
German      & 18.24   & 30.23   & --      & 32.05   & 3.28    & --      & 20.19   & 17.87   & --      \\
Greek       & --      & 30.96   & --      & --      & --      & --      & --      & --      & --      \\
Hebrew      & --      & --      & --      & --      & --      & 1.54    & --      & --      & --      \\
Hindi       & --      & --      & --      & --      & --      & --      & --      & 7.92    & --      \\
Indonesian  & --      & 31.04   & --      & --      & --      & --      & --      & --      & --      \\
Italian     & --      & --      & --      & --      & --      & --      & 23.48   & --      & --      \\
Japanese    & --      & --      & --      & --      & --      & --      & 16.48   & --      & 12.86      \\
Kannada     & --      & --      & --      & --      & --      & 1.39    & --      & --      & --      \\
Polish      & --      & --      & --      & --      & --      & --      & 19.26   & --      & --      \\
Portuguese  & 17.57   & --      & --      & --      & --      & --      & --      & --      & --      \\
Russian     & --      & --      & --      & --      & --      & --      & --      & --      & 9.57 \\
Spanish     & 15.54   & --      & --      & --      & --      & --      & 21.11   & 16.73   & 2.64      \\
Swedish     & --      & 31.77   & --      & --      & --      & --      & --      & --      & --      \\
Thai        & --      & 23.74   & --      & --      & --      & --      & --      & 12.13   & --      \\
Turkish     & --      & --      & --      & --      & --      & --      & 20.80   & --      & --      \\
Vietnamese  & --      & --      & 27.95   & --      & --      & --      & --      & --      & --      \\
\bottomrule
\end{tabular}
}
\caption{The performance of Codex with Cross-lingual Zero-Shot Transfer setting on different datasets and different languages.}
\label{tab:codex_crosszero}
\end{table*}

\begin{table*}[ht!]
\centering
\resizebox{\textwidth}{!}{
\begin{tabular}{@{}llllllllll@{}}
\toprule
            & MATIS & MGeoQuery & MSpider & MNLmaps & MOvernight & MCWQ & MSchema2QA & MTOP & MCoNaLa \\\midrule
Arabic      & --      & --      & --      & --      & --      & --      & 5.66    & --      & --      \\
Chinese     & 0.00    & 16.07   & 2.61    & --      & 0.47    & 0.00    & 4.63    & --      & --      \\
Farsi       & --      & 3.34    & --      & --      & --      & --      & 1.54    & --      & --      \\
Finnish     & --      & --      & --      & --      & --      & --      & 1.13    & --      & --      \\
French      & 0.00    & --      & --      & --      & --      & --      & --      & 1.54    & --      \\
German      & 0.00    & 16.43   & --      & 7.05    & 0.29    & --      & 6.49    & 1.94    & --      \\
Greek       & --      & 9.84    & --      & --      & --      & --      & --      & --      & --      \\
Hebrew      & --      & --      & --      & --      & --      & 0.00    & --      & --      & --      \\
Hindi       & --      & --      & --      & --      & --      & --      & --      & 1.78    & --      \\
Indonesian  & --      & 18.50   & --      & --      & --      & --      & --      & --      & --      \\
Italian     & --      & --      & --      & --      & --      & --      & 5.66    & --      & --      \\
Japanese    & --      & --      & --      & --      & --      & --      & 2.37    & --      & 0.08      \\
Kannada     & --      & --      & --      & --      & --      & 0.00    & --      & --      & --      \\
Polish      & --      & --      & --      & --      & --      & --      & 3.71    & --      & --      \\
Portuguese  & 0.00    & --      & --      & --      & --      & --      & --      & --      & --      \\
Russian     & --      & --      & --      & --      & --      & --      & --      & --      & 0.09 \\
Spanish     & 0.00    & --      & --      & --      & --      & --      & 7.83    & 2.26    & 0.04      \\
Swedish     & --      & 14.62   & --      & --      & --      & --      & --      & --      & --      \\
Thai        & --      & 0.27    & --      & --      & --      & --      & --      & 0.81    & --      \\
Turkish     & --      & --      & --      & --      & --      & --      & 0.31    & --      & --      \\
Vietnamese  & --      & --      & 0.79    & --      & --      & --      & --      & --      & --      \\
\bottomrule

\end{tabular}
}
\caption{The performance of BLOOM with Cross-lingual Zero-Shot Transfer setting on different datasets and different languages.}
\label{tab:bloom_crosszero}
\end{table*}

\begin{table*}[ht!]
\centering
\resizebox{.9\textwidth}{!}{
\begin{tabular}{@{}lllllllll@{}}
\toprule
            & ATIS & GeoQuery & Spider & NLmaps & Overnight & MCWQ-MCD3 & Schema2QA & MTOP \\\midrule
Arabic      & --      & --      & --      & --      & --      & --      & 53.66   & --      \\
Chinese     & 4.16    & 23.22   & 44.12   & --      & 46.61   & 14.35   & 37.49   & --      \\
Farsi       & --      & 29.00   & --      & --      & --      & --      & 46.55   & --      \\
Finnish     & --      & --      & --      & --      & --      & --      & 57.16   & --      \\
French      & 24.40   & --      & --      & --      & --      & --      & --      & 75.10   \\
German      & 23.27   & 65.31   & --      & 64.89   & 57.44   & --      & 61.77   & 73.81   \\
Greek       & --      & 70.91   & --      & --      & --      & --      & --      & --      \\
Hebrew      & --      & --      & --      & --      & --      & 22.53   & --      & --      \\
Hindi       & --      & --      & --      & --      & --      & --      & --      & 72.35   \\
Indonesian  & --      & 71.90   & --      & --      & --      & --      & --      & --      \\
Italian     & --      & --      & --      & --      & --      & --      & 58.29   & --      \\
Japanese    & --      & --      & --      & --      & --      & --      & 39.79   & --      \\
Kannada     & --      & --      & --      & --      & --      & 23.61   & --      & --      \\
Polish      & --      & --      & --      & --      & --      & --      & 53.45   & --      \\
Portuguese  & 23.27   & --      & --      & --      & --      & --      & --      & --      \\
Spanish     & 3.46    & --      & --      & --      & --      & --      & 63.72   & 78.33   \\
Swedish     & --      & 68.38   & --      & --      & --      & --      & --      & --      \\
Thai        & --      & 28.82   & --      & --      & --      & --      & --      & 64.35   \\
Turkish     & --      & --      & --      & --      & --      & --      & 63.23   & --      \\
Vietnamese  & --      & --      & 43.24   & --      & --      & --      & --      & --      \\
\bottomrule
\end{tabular}
}
\caption{The performance of XLM-R+PTR with Cross-lingual Few-Shot Transfer setting on different datasets and different languages.}
\label{tab:xlmr_crossfew}
\end{table*}

\begin{table*}[ht!]
\centering
\resizebox{.9\textwidth}{!}{
\begin{tabular}{@{}lllllllll@{}}
\toprule
            & MATIS & MGeoQuery & MSpider & MNLmaps & MOvernight & MCWQ & MSchema2QA & MTOP \\\midrule
Arabic      & --      & --      & --      & --      & --      & --      & 47.89   & --      \\
Chinese     & 48.65   & 44.32   & 44.39   & --      & 60.40   & 29.48   & 53.35   & --      \\
Farsi       & --      & 44.23   & --      & --      & --      & --      & 42.22   & --      \\
Finnish     & --      & --      & --      & --      & --      & --      & 61.48   & --      \\
French      & 50.45   & --      & --      & --      & --      & --      & --      & 62.81   \\
German      & 50.32   & 56.95   & --      & 71.70   & 64.67   & --      & 68.80   & 80.68   \\
Greek       & --      & 60.11   & --      & --      & --      & --      & --      & --      \\
Hebrew      & --      & --      & --      & --      & --      & 26.85   & --      & --      \\
Indonesian  & --      & 58.40   & --      & --      & --      & --      & --      & --      \\
Italian     & --      & --      & --      & --      & --      & --      & 66.63   & --      \\
Japanese    & --      & --      & --      & --      & --      & --      & 45.73   & --      \\
Kannada     & --      & --      & --      & --      & --      & 18.21   & --      & --      \\
Polish      & --      & --      & --      & --      & --      & --      & 57.98   & --      \\
Portuguese  & 49.32   & --      & --      & --      & --      & --      & --      & --      \\
Spanish     & 49.10   & --      & --      & --      & --      & --      & 65.81   & 83.51   \\
Swedish     & --      & 64.71   & --      & --      & --      & --      & --      & --      \\
Thai        & --      & 44.49   & --      & --      & --      & --      & --      & 71.71   \\
Turkish     & --      & --      & --      & --      & --      & --      & 69.00   & --      \\
Vietnamese  & --      & --      & 54.45   & --      & --      & --      & --      & --      \\
\bottomrule
\end{tabular}
}
\caption{The performance of mT5 with Cross-lingual Few-Shot Transfer setting on different datasets and different languages.}
\label{tab:mt5_crossfew}
\end{table*}

\end{document}